\documentclass[11pt, a4paper, logo, onecolumn, copyright]{googledeepmind}

\usepackage[authoryear, compress, round]{natbib}
\usepackage{xspace}  
\usepackage{siunitx}  
\usepackage[capitalise,noabbrev]{cleveref}
\usepackage{placeins}
\usepackage{subfig}
\usepackage{booktabs}   
\usepackage{tabularx}   
\usepackage{amsmath}    
\usepackage{multirow}
\usepackage{url}
\usepackage{lineno}
\usepackage{hyperref}

\usepackage[textsize=tiny,color=orange!40]{todonotes}

\title{WeatherNext 3: Increasing resolution and performance of global weather models with raw observations}

\author[1*]{Stephan Rasp}
\author[2*]{Boris Babenko}
\author[1*]{Dominic Masters}
\author[1*]{Andrew El-Kadi}
\author[2*]{Samier Merchant}
\author[1*]{Guy Shalev}
\author[1*]{Ilan Price}
\author[2]{Fred Zyda}
\author[1]{Remi Lam}
\author[1]{Sasha Shysheya}
\author[1]{Matthew Willson}
\author[1]{Stratis Markou}
\author[1]{Shreya Agrawal}
\author[1]{Suhani Vora}
\author[1]{Mohammed Alewi Hassen}
\author[1]{Sunny Mak}
\author[1]{Tom R. Andersson}
\author[3]{Megan Bela}
\author[2]{Akib Uddin}
\author[1]{Nofar Peled Levi}
\author[1]{Ben Gaiarin}
\author[1*]{Ferran Alet}
\author[2]{Aaron Bell}
\author[1]{Peter Battaglia}
\author[1*]{Alvaro Sanchez-Gonzalez}
\affil[*]{Equal contribution}
\affil[1]{Google DeepMind}
\affil[2]{Google Research}
\affil[3]{Google}

\correspondingauthor{{\{srasp, bbabenko, dominicmasters, aelkadi, alvarosg\}@google.com}}

\newcommand{\twentiethdegree}{\ang{0.05}\xspace}
\newcommand{\tenthdegree}{\ang{0.1}\xspace}

\newcommand{\quarterdegree}{\ang{0.25}\xspace}
\newcommand{\halfdegree}{\ang{0.5}\xspace}

\newcommand{\onedegree}{\ang{1}\xspace}
\newcommand{\hresfczero}{HRES-fc0\xspace}
\newcommand{\hresfc}{HRES-fc0-5\xspace}

\begin{abstract}
State-of-the-art AI weather models have shown impressive medium-range forecast skill and computational efficiency, but suffer two key shortcomings: their forecasts have lower spatial and temporal resolution than the best physics-based models and they are exclusively initialized with and trained on analysis data. As a result, they cannot directly make use of observations, and any biases in the analysis are inherited by the forecast. 
WeatherNext 3 addresses these shortcomings and establishes a new state-of-the-art for probabilistic medium-range forecasting skill. 
First, WeatherNext 3 generates new forecasts every hour (rather than every 6 hours like traditional global models) by ingesting low-latency geostationary satellite data.
Second, WeatherNext 3's temporal and spatial resolution are on par with physics-based global models, with hourly time steps and \tenthdegree resolution for single-level variables, including solar radiation and cloud cover.
Third, WeatherNext 3 moves beyond traditional analysis variables by learning to predict satellite-derived precipitation estimates, as well as tropical cyclone and station observations.
Modelling sparse station data allows WeatherNext 3 to make 2m temperature and dewpoint predictions at any location and time, conditioned on local geographical features, with substantially lower error than competing global models, even when evaluated against unseen stations.
Together, WeatherNext 3's capabilities move operational AI-based weather forecasting beyond emulating the traditionally distinct stages of data assimilation, forecasting and post-processing, which helps to further push the frontier of performance and granularity for global weather prediction.

\end{abstract} 

\begin{document}

\maketitle

\section{Introduction}

AI weather models are revolutionizing weather forecasting, surpassing traditional models on many measures of skill, while also being faster and cheaper to run \citep{lam2023learning,bi2023accurate,rasp2024weatherbench,price2025probabilistic,kossaifi2026demystifying}. Several AI models now run operationally and are gaining widespread adoption \citep{alet2026operational, lang2026aifs}. So far, most state-of-the-art global AI models have been initialized with and trained against (re-)analysis data, which are gridded estimates of the state of the atmosphere produced by assimilating a wide range of observations using a traditional numerical weather prediction (NWP) model \citep{hersbach2020era5, soci2024era5}.

Analyses have proven very effective for training AI models, but have shortcomings that are inherited by models trained exclusively on this data. First, analyses have known biases, particularly for key variables like precipitation or surface temperature \citep{lavers2022evaluation, liu2024global}. Second, global operational analyses are only produced every 6 hours, assimilating observations up to 3 hours past the analysis date. Combined with a 3 hour latency in producing and publishing the analysis, this means that the most recent analysis state is always between 6 and 12 hours stale \citep{ecmwf_dissemination_schedule}. Third, some observations, such as those coming from geostationary satellites, are only indirectly assimilated \citep{lean2023using}, resulting in a loss of information.  

\begin{figure}[h!]
    \centering\includegraphics[width=1.0\linewidth]{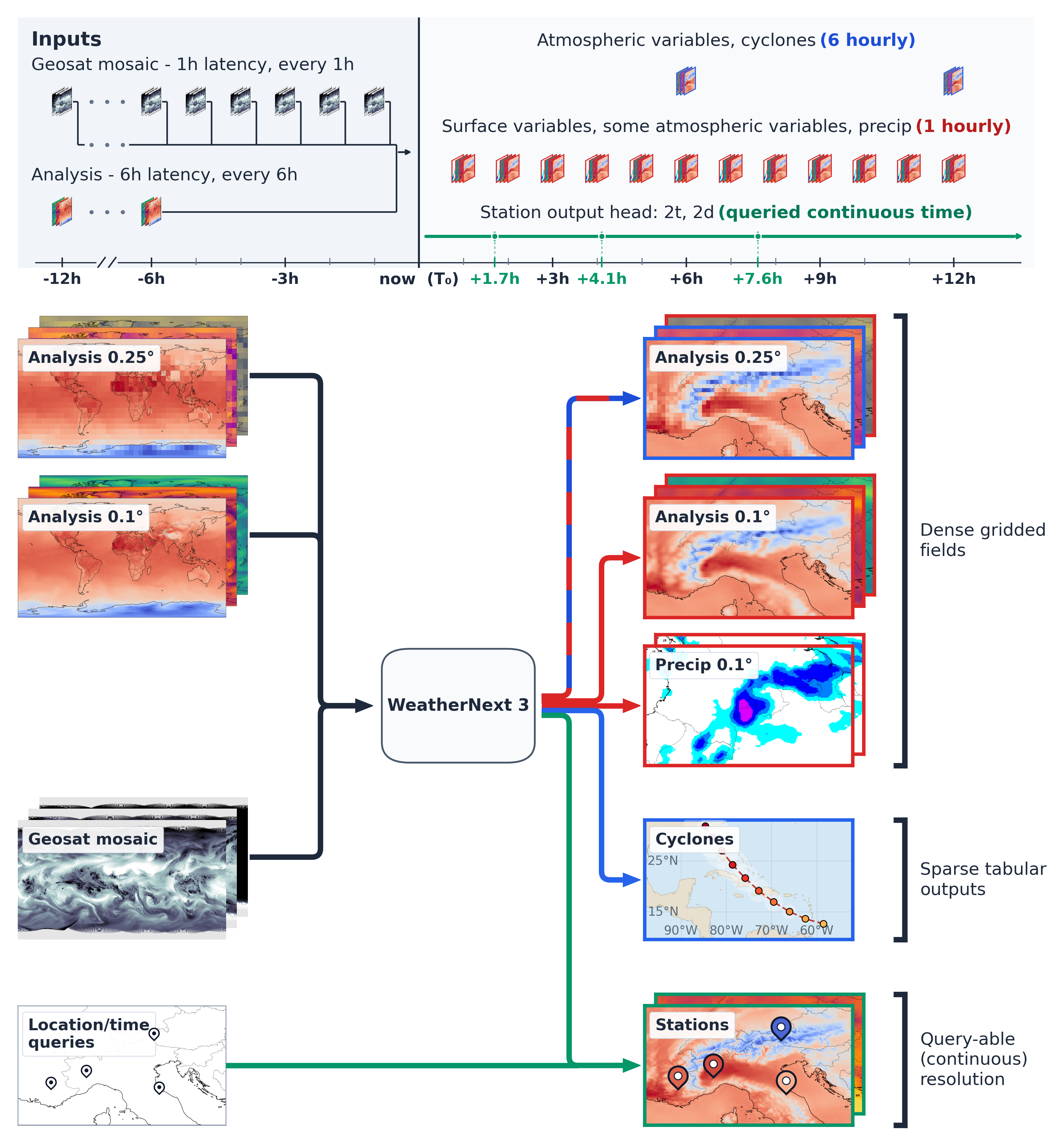}
    \caption{\textbf{Model overview.} Schematic showing the different input and output modalities, and spatiotemporal resolutions of WeatherNext 3 (WN3). WN3 takes two sources of input: (a) two analysis frames (6 hours apart), with an operational latency of 5 hours, and (b) the 12 most recent geostationary satellite mosaic frames, which are available every hour and the most recent of which has an operational latency of just under 1 hour.  Analysis inputs are \tenthdegree resolution for a subset of key atmospheric variables and surface variables, and \quarterdegree for other variables. WN3 predicts precipitation, surface variables and some atmospheric variables at hourly temporal resolution, with the remaining variables predicted in 6 hour timesteps. WN3's station output head predicts 2m temperature and 2m relative humidity at arbitrary spatial and temporal resolution (i.e. is continuously query-able in space and time). WN3 also produces off-grid sparse or tabular cyclone predictions at 6-hourly temporal resolution.~\looseness=-1
    }
    \label{fig:figure_1}
\end{figure}

There has been a growing interest in directly leveraging observational data in AI weather models to address these shortcomings. This was first done in regional models that ingest and predict observational data directly \citep{sonderby2020metnet, ravuri2021skilful, andrychowicz2023deep, zhao2024omg, zhao2026skillful}. Such models have proven particularly successful at precipitation nowcasting but, because of their limited spatial context, are restricted to short-range prediction. More recently, there has also been emerging research into AI-based global data assimilation, either going from observations to analysis \citep{vaughan2024aardvark, ni2025huracan, gupta2026healda, windborne2026weathermesh6}, or directly from observations to predicted observations \citep{pinnington2026aifs}. Results are promising, but so far most global models are not operational or have a lower spatial resolution than traditional NWP systems. We recently published WeatherNext-Cyclones \citep{alet2026operational}, which combines training on global analysis with observation cyclone data, resulting in a new state-of-the-art in operational tropical cyclone prediction during the 2025 hurricane season~\citep{cangialosi2026national}.

WeatherNext 3 (WN3) further builds on these ideas by directly ingesting and predicting raw observations, setting a new state-of-the-art on analysis and observation benchmarks. Specifically:

\begin{itemize}
    \item WN3 produces a new forecast every hour, ingesting the latest geostationary satellite imagery.
    \item WN3 produces hourly, \tenthdegree output of all single-level variables including variables that were not available in WeatherNext 2 (WN2), such as solar radiation and cloud cover.
    \item WN3 is trained to predict global hourly precipitation products derived from satellite data.
    \item WN3's station head makes geographically-informed predictions of 2m temperature and dewpoint at any location on the globe, and any point in time, as opposed to predefined grid points and time steps in traditional models.
    \item WN3 substantially improves the \quarterdegree analysis and cyclone specific forecasts of WeatherNext 2 Cyclones \citep{alet2026operational}.
\end{itemize}
WeatherNext 3 is running operationally. For more information on how to access the forecasts, see \url{https://deepmind.google/science/weathernext/}.


\section{WeatherNext 3}
\label{sec:wn3}

\subsection{Problem formulation.}
The weather forecasting problem we address is to learn a distribution over future weather trajectories conditional on analysis and recent observations. The atmosphere at time $t$ and spatial location $s$ is represented by a set of $N$ spatio-temporal fields (indexed by $i$),
\begin{equation}
\label{eq:state}
\mathbf{W}_t = \bigl\{ f_i(s, t) : \mathcal{S}_i \times \mathcal{T}_i \to \mathcal{V}_i \bigr\}_{i=1}^{N},
\end{equation}
where each field $f_i$ denotes a physical variable over a continuous or discrete spatial domain $\mathcal{S}_i \subseteq \mathbb{S}^2$, continuous or discrete time subset $\mathcal{T}_i \subseteq \mathbb{R}$, and value space $\mathcal{V}_i$. Whereas WN2 was exclusively modelling a single data modality, namely analysis variables at \quarterdegree, WN3 has several modality groups, within which individual variables and vertical levels are separate fields sharing a common spatial grid and temporal cadence (\cref{tab:data}).

We seek to learn the conditional distribution
\begin{equation}
\label{eq:trajectory}
p\!\left(\mathbf{W}_{(0,\,T]} \;\middle|\; \mathbf{W}_{\le 0}\right)
\end{equation}
over weather trajectories from the present to lead time~$T$, where subscripts denote continuous-time intervals. We discretise the forecast into $T/\Delta$ outer steps of width $\Delta = 6\,\text{h}$. Modelling the atmosphere as a 2nd-order Markov process, the trajectory distribution factorises as
\begin{equation}
\label{eq:markov}
p\!\left(\mathbf{W}_{(0,\,T]} \;\middle|\; \mathbf{W}_{\le 0}\right) = \prod_{t=0}^{T/\Delta-1} p\!\left(\mathbf{W}_{(t\Delta,\;(t+1)\Delta]} \;\middle|\; \mathbf{W}_{((t{-}2)\Delta,\;(t{-}1)\Delta] }, \mathbf{W}_{((t{-}1)\Delta,\;t\Delta]}\right).
\end{equation}
Each factor predicts the weather over a $\Delta$-wide window given the two preceding windows. Because some fields~$f_i$ have temporal supports finer than~$\Delta$ (e.g. hourly single-level analysis fields or precipitation estimates), the model simultaneously predicts different weather time steps within each window by treating each inner time step as its own variable.

Some of the fields are modelled as autoregressive: these are fed back as input to the next step. The remainder are \emph{target-only}: predicted but not re-ingested (\cref{tab:data}). This detail is omitted from the simplified mathematical formulation above to avoid clutter in the exposition. 

\subsection{Key features.}

WN3 is based on an encode-process-decode architecture and leverages the Functional Generative Networks (FGN) approach which already underpins WN2 \citep{alet2025skillful}. It produces 15-day 64-member ensemble forecasts, where individual members are sampled by injecting noise into the model. The training objective is to minimize the marginal (i.e. per-modality, per-level, per-location and per-time step) continuous ranked probability score (CRPS). WN2 used this approach to achieve state-of-the-art probabilistic medium-range weather prediction at \quarterdegree spatial and 6 hourly temporal resolution. ECMWF's AIFS ENS \citep{lang2026aifs} uses a related approach with some architectural and loss differences. 
WN3 extends this framework and scales up the training with respect to WN2 (see \cref{fig:figure_1}). This section describes the key features of WN3; for details refer to \cref{sec:app_model}.

\paragraph{Multiple data modalities and resolutions.}
WN3 natively handles data modalities with different spatial resolutions, temporal cadences and operational latencies. Architecturally, different spatial resolutions (such as \quarterdegree and \tenthdegree grids) are handled by separate encoders and decoders that map the data from its native resolution to the shared icosahedral processor mesh and back again. This design allows each modality to interact at its natural resolution without requiring all data to be regridded to a common grid. At the same time, it also enables dynamics to be propagated on the processor mesh independently of the native data resolution or data availability.\\
To fill in the intermediate hourly frames of the 6 hourly outputs, WN2 learned a separate $6\text{h}\rightarrow1\text{h}$ upsampler model with the same architecture but different parameters
\footnote{https://developers.google.com/weathernext/guides/model-specs-vmg}. In contrast, WN3 predicts hourly temporal resolutions directly, making it a native feature of the model.

\paragraph{Observational input and hourly initialization.}
Leveraging its multimodal capabilities, WN3 now ingests geostationary satellite data directly, in addition to analysis. Because satellite data is available more frequently than the 6 hourly HRES analysis, we can generate a new forecast every hour (\cref{subsection:Data}). We find that this rapid refresh yields skill improvements at short lead times, particularly for fast-evolving variables like precipitation (\cref{sec:precip_eval}).

\begin{table}[!htb]
\centering
\scriptsize
\setlength{\tabcolsep}{5pt}
\begin{tabularx}{\textwidth}{@{} >{\raggedright\arraybackslash}p{2.2cm} >{\raggedright\arraybackslash}X p{2.4cm} >{\raggedright\arraybackslash}p{2.2cm} >{\raggedright\arraybackslash}p{3.4cm} @{}}
\toprule
\textbf{Dataset} & \textbf{Variables} & \textbf{Time period} & \textbf{Spatial resolution} & \textbf{Temporal resolution} \\
\midrule
\multicolumn{5}{@{}l}{\textbf{Inputs}} \\
\midrule
ERA5 & 13 pressure levels: \textit{z}, \textit{t}, \textit{q}, \textit{u}, \textit{v}, \textit{w} \newline Single-level: \textit{2t}, \textit{2d}, \textit{10u}, \textit{10v}, \textit{100u}, \textit{100v}, \textit{msl}, \textit{sst}, \textbf{\textit{tcc}}, \textbf{\textit{hcc}}, \textbf{\textit{mcc}}, \textbf{\textit{lcc}} & \textbf{1959}-2015 & \quarterdegree & 6h \\
\addlinespace[4pt]
\hresfc & Same as ERA5 & 2016-2023/\textbf{26*} & \quarterdegree + \textbf{\tenthdegree (Single-level variables)} & 6h \\
\addlinespace[4pt]
\textbf{Geostationary satellite mosaic} & 11 channels & 2016-2023/\textbf{26*} & \tenthdegree & 1h \\
\addlinespace[4pt]
\textbf{Sparse decoder metadata} & Elevation, land/sea mask & static & sparse training / \twentiethdegree inference & static \\
\addlinespace[4pt]
Forcings & Elapsed year progress, local time of day, \textbf{tisr} & deterministic &  global + \quarterdegree + \textbf{\tenthdegree} & 6h + \textbf{1h} (tisr)   \\
\addlinespace[4pt]
  & Land/sea mask, geopotential at surface & static &  \quarterdegree + \textbf{\tenthdegree} & static   \\
\midrule
\multicolumn{5}{@{}l}{\textbf{Targets}} \\
\midrule
ERA5 & 13 pressure levels: \textit{z}, \textit{t}, \textit{q}, \textit{u}, \textit{v}, \textit{w} \newline Single-level: \textit{2t}, \textit{2d}, \textit{10u}, \textit{10v}, \textit{100u}, \textit{100v}, \textit{msl}, \textit{sst}, \textbf{\textit{tcc}}, \textbf{\textit{hcc}}, \textbf{\textit{mcc}}, \textbf{\textit{lcc}}, \textbf{\textit{cdir}}$^\dagger$, \textbf{\textit{fdir}}, \textbf{\textit{ssrd}}, \textit{tp} & \textbf{1959}-2015 & \quarterdegree & 6h + \textbf{1h} (single-level variables; 300/500 hPa \textit{t}/\textit{z}; 1000hPa \textit{u}/\textit{v}) \\
\addlinespace[4pt]
\hresfc & Same as ERA5 & 2016-2023/\textbf{26*} & \quarterdegree + \textbf{\tenthdegree (single-level variables)} & 6h + \textbf{1h} (single-level variables; 300/500 hPa \textit{t}/\textit{z}; 1000hPa \textit{u}/\textit{v}) \\
\addlinespace[4pt]
\textbf{Geostationary satellite mosaic} & 11 channels & 2016-2023/\textbf{26*} & \tenthdegree & 1h \\
\addlinespace[4pt]
\textbf{PARDIG} & 1h precipitation accumulation & 1999-2023/\textbf{26*} & \tenthdegree & 1h \\
\addlinespace[4pt]
\textbf{IMERG} & 1h precipitation accumulation & 2000-2023/\textbf{25*} & \tenthdegree & 1h \\
\addlinespace[4pt]
\textbf{Surface observations} & \textit{2t}, \textit{2d} & 2016-2023/\textbf{26*} & sparse training / \twentiethdegree inference & 1h \\
\addlinespace[4pt]
Cyclone data & Existence, minimum pressure, maximum wind, radius of max wind, wind quadrant radii at three wind speeds & 1979-2023/\textbf{26*} & \quarterdegree $\rightarrow$ \text{table} & 6h \\
\bottomrule
\end{tabularx}
\caption{Table of model inputs and targets. See \cref{tab:var_abbrev_single} for a list of variable abbreviations. Newly added features with respect to WN2 highlighted in \textbf{bold}. (*) Model trained until the end of 2023 for 2024 evaluations. Production model trained until June 30 2026. ($\dagger$) Note that while WN3 was trained to output \textit{cdir} as one of its output variables, during development it was observed that skill for this variable was an outlier and substantially worse than expected. We thus do not make WN3 \textit{cdir} forecasts available operationally, and so do not include evaluations thereof in this paper.}
\label{tab:data}
\end{table}

\paragraph{Continuous output in time and space via latent interpolation.}
To model sparse, i.e. non-gridded data sources, such as in-situ surface observations, we decode predictions as a continuous query. This is done by interpolating the \tenthdegree latent to the specific sparse observation location and combining it with relevant metadata, namely the exact station elevation, whether it is a land (station) or sea (buoy or ship) observation and the relative time offset to the outer 6 hour time step (\cref{sec:app_model}), before making the 2m temperature and dewpoint predictions. Since the output head is trained globally and metadata can easily be obtained for any location on Earth, this allows WN3 to make predictions at any arbitrary location and time at inference. In production inference and for evaluation in this paper, we create predictions at a \twentiethdegree grid and at hourly intervals. Technically, it would be possible to make predictions at precise station locations, however, in practice this only made a marginal difference in our evaluations.

\paragraph{Scaling model size and multi-stage training curriculum.}
WN3 increases the latent size from 768 to 1024 and the mesh transformer depth from 24 to 32 layers with respect to WN2. To fit this larger model on available accelerators, we introduce spatial sharding of the processor mesh and grid-to-mesh and mesh-to-grid  gather/scatter operations (\cref{sec:app_model}). Further, we modify the training curriculum by progressively increasing resolution from \onedegree to \quarterdegree to \tenthdegree, followed by a frozen finetuning stage for the sparse station head. We also make use of ERA5 dating back to 1959 (from 1979 for WN2) and all of its hourly initializations. See \cref{sec:app_training} for a detailed list of training stages including which data modalities are added at which stage.

\paragraph{Epistemic dropout.}
WN3 uses functional perturbation ensembles \citep{alet2025skillful} to capture both aleatoric and epistemic uncertainty. Aleatoric uncertainty is modelled by perturbing the normalisation layers with a low-dimensional noise vector. In WN2, epistemic uncertainty was modelled with deep ensembles by training 4 independent model seeds. WN3 only uses 2 seeds, but adds \emph{epistemic dropout}.

Dropout has been successfully used as a model perturbation during CRPS training to capture aleatoric uncertainty \citep{cachay2026u, diaconu2026otter}. Epistemic dropout instead builds on the framing of dropout as sampling from an exponential family of networks \citep{gal2016dropout}, and combines it naturally with our aleatoric uncertainty training. During inference, the dropout mask is independently sampled for each ensemble member and time step. During training, however, a single dropout mask is sampled and shared across ensemble members for that step. This prevents the samples from exploiting dropout variance, and instead effectively trains one of the dropout subnetworks. This naturally expands the spread of evaluation samples relative to training, compensating for the fact that the model learns to calibrate its ensemble spread based on its training skill, which in practice is better than its evaluation skill. Given the increased model capacity and smaller number of model seeds, this proved important in preserving ensemble spread.

\subsection{Model inputs and targets}
\label{subsection:Data}

A list of all data used in WN3 is provided in \cref{tab:data}.

\paragraph{Atmospheric analysis.}
We use a combination of ERA5 \citep{hersbach2020era5} and HRES \citep{ecmwf_hres_2023} for training WN3. In prior work, we used \hresfczero which is available 6 hourly. For hourly initialization and hourly targets we use the most recent HRES forecasts in-between the 6 hour intervals. This is denoted as \hresfc throughout the text. For the \onedegree and \quarterdegree training stages, we use ERA5 from 1959 to 2015, and \hresfc from 2016 onward. For the \tenthdegree variables, we train exclusively on \hresfc starting in 2016 (details in \cref{sec:app_training}). Note that the model treats the \quarterdegree analysis (3D and single-level fields) and the \tenthdegree analysis (only single-level fields) as separate modalities. Single-level variables are predicted at hourly temporal resolution as separate channels with time offsets \{-5, -4, -3, -2, -1\}\,\text{h} relative to the 6-hourly analysis prediction. In addition, some pressure level \quarterdegree variables (300/500 hPa \textit{t}/\textit{z}; 1000hPa \textit{u}/\textit{v}) are also predicted at hourly resolution.

For the model used throughout most of the evaluations in this paper, we train until the end of 2023. For the production model, we train until June 30 2026. ECMWF released a new model cycle (50r1) on May 12 2026\footnote{https://www.ecmwf.int/en/newsletter/185/earth-system-science/upgrade-ifs-cycle-50r1}. To provide WN3 with as much data of the new cycle as possible, we also use the experimental 50r1 data available from January 20 to May 11 2026.

\paragraph{Geostationary satellite mosaics.}
WN3 uses an 11-channel \tenthdegree hourly mosaic of geostationary satellite imagery produced by combining data streams of several satellites (\cref{sec:app_data}). Geostationary satellites cover the infrared and visible spectrum, providing important information on water vapor, temperature and the location of clouds. Our mosaic has an operational latency of roughly 1 hour, giving us data which is at least 5 hours more recent than the most recent analysis. To exploit this we feed geostationary satellite data with time offsets \{0, 1, 2, 3, 4, 5\}\,h with respect to the analysis state to the model. In order for the model to be able to propagate this information over many rollout steps, we predict the geostationary channels autoregressively.

As an example of how these time offsets play out in practice, at 14 UTC, the latest geostationary mosaic is from 13 UTC, so the model input contains data from \{08, 09, 10, 11, 12, 13\}\,UTC. The latest available HRES analysis at this point is from 06 UTC, so we input the 2h HRES forecast, valid at 08 UTC.~\looseness=-1

\paragraph{Precipitation estimates.}
In addition to analysis precipitation, which was the sole precipitation product predicted by WN2, we added two additional, more specialized precipitation products to WN3: a novel experimental precipitation estimate, called PARDIG, that is based on a separate AI model trained to predict sparse space-borne radar data from the GPM core observatory\footnote{https://gpm.nasa.gov/missions/GPM/core-observatory}; and the 'Final' version of NASA's Integrated Multi-satellitE Retrievals for GPM (IMERG), a widely used global precipitation product \citep{huffman2023nasa}. Our internal analysis shows that PARDIG matches independent observations from ground-based radar and rain gauges more closely than IMERG. Nevertheless, due to the popularity of IMERG we include IMERG-based predictions as well. Both products have a spatial resolution of \tenthdegree. PARDIG is available from 1999, while IMERG availability starts in June 2000 (currently IMERG Final is only available until September 2025). See \cref{sec:app_data} for a detailed description of both products.

\paragraph{In-situ surface observations.}
We combine three surface observation datasets for model training, starting in 2001: METAR and Mesonet (both from MADIS\footnote{https://madis.ncep.noaa.gov/}) and ICOADS \citep{freeman2017icoads}. METAR is a set of roughly 5,000 airport weather stations, while Mesonet is a collection of high-density regional station networks, primarily in Europe and North America. The number of Mesonet stations changes with time. In 2024, it was roughly 15,000. ICOADS is a dataset of ship and buoy measurements. The quantity of observations fluctuates greatly over time but, on average in 2024, they number around 3000 per hour. See \cref{sec:app_data} for more details. We hold out a random but temporally consistent 5\% of METAR and Mesonet stations from training, which we can then use for independent evaluation of spatial generalization. The station head is trained to predict 2m temperature and dewpoint. 10m wind speed observations are used for evaluation only.

\cref{fig:fig_a1_station_distributions} highlights the substantial gaps in spatial coverage of surface in-situ observations. As a result we noticed strong biases in our model predictions in sparsely observed, out-of-distribution areas, such as the Andes or Himalayas, and some high-latitude ocean regions. To alleviate these issues, we add "pseudo-station" data by randomly sampling 2,000 locations per hour and filling these with interpolated values from the corresponding variable in ERA5 or \hresfc. This did not result in a degradation of the predictions when evaluated against hold-out stations but greatly reduced biases.

\paragraph{Tropical cyclones.} Our model directly learns to predict the existence, position and attributes of tropical cyclones, as done in \citet{alet2026operational}. We similarly convert centre positions, maximum sustained wind speeds~$V_{\text{max}}$, and 34/50/64-knot wind radii from the International Best Track Archive \citep[IBTrACS,][]{knapp2010ibtracs} into discrete gridded targets. We make only very minor modifications to this process, documented in Appendix~\ref{app:cyclones}. Additionally, we interpolate IBTrACS to hourly resolution before griddification to obtain cyclone targets for training with hourly initializations.

\section{Evaluation methodology}
Our evaluation philosophy is to identify the current state-of-the-art model for each task and to use this as a primary baseline for our scorecards. This follows our previous work \citep{price2025probabilistic, alet2025skillful, alet2026operational}. Since WN3 has an extended set of capabilities, no single baseline model covers all evaluations. Similarly, for each task we identify the appropriate independent ground truth(s) against which to evaluate. When dealing with different forecast resolutions, we interpolate lower resolution forecasts to the ground truth resolution. While this introduces some subtleties, we believe that this ultimately is the best representation of actual forecast skill. Evaluations are done for the full year of 2024, using a model trained until the end of 2023, except for the real-time evaluation, which uses a model trained until June 30 2026. Aside from the latency-adjusted evaluation, all reported lead times are relative to nominal initialization (e.g. for WN3 this corresponds to the nominal time of the input analysis).

\subsection{Ground truths}
\label{sec:gt}
\paragraph{Analysis.}
To evaluate the \quarterdegree and \tenthdegree analysis predictions we use \hresfczero as ground truth, which is also what WN3 is initialized from. One exception is the evaluation of AIFS ENS, for which we use the AIFS ENS control fc0 as ground truth, which derives from the same raw data as \hresfczero but has differences in interpolation (see \cref{sec:app_data} for details). We note that ECMWF typically evaluates against HRES-an, which involves an additional post-processing step that mainly affects surface temperature. Evaluating directly against weather stations as an independent ground truth allows us to bypass this ambiguity. Given the absence of hourly \hresfczero and HRES-an, we can only evaluate WN3's hourly analysis forecasts against \hresfc, which is done for $\leq2$ day lead times in \cref{fig:fig_appendix_0p1_analysis_scores_48h}.

\paragraph{Weather stations.}
We use the set of METAR and Mesonet stations not seen during training for independent evaluation. To be consistent, this also applies to wind speed measurements, which were not explicitly used for training. We use bilinear interpolation to go from model grid to station location. This also applies to the WN3 station head predictions which were queried at a \twentiethdegree grid and then interpolated. Using the nearest grid point led to worse results, with bigger differences for the coarser forecasts. Following ECMWF, for 2m temperature we apply a lapse rate adjustment to account for differences between the model grid and station elevation \citep{ingleby2015global}. However we only apply this to \quarterdegree and \tenthdegree forecasts. For our \twentiethdegree station head predictions, we only saw a negligible effect on metrics and decided to show the raw scores. We also evaluate relative humidity, which is computed from the temperature and dewpoint forecasts and observations.

\paragraph{Precipitation.}
We evaluate against three precipitation ground truths: IMERG Final, the  Multi Radar Multi Sensor (MRMS) Quantitative Precipitation Estimate Pass-2 product \citep{smith2016multi} and rain gauges. Evaluating against a range of independent precipitation products is important since no single product is perfect and they disagree a fair amount \citep{moazami2021comprehensive}. By choosing three datasets with different trade-offs, we aim to provide a representative picture of precipitation forecast performance.

IMERG provides a precipitation estimate based on data from the GPM satellite constellation. It has the benefit of being available globally at \tenthdegree resolution. However, for WN3 it is not an independent ground truth since IMERG is used as a model target. This naturally applies to the IMERG output head but could also indirectly influence the PARDIG head since both are co-trained end-to-end. Note that IMERG Final is unavailable for the real-time evaluation. 

MRMS is primarily based on ground radar in the contiguous United States. The Pass 2 precipitation product is additionally calibrated against rain gauge measurements. MRMS is considered a high-quality dataset, more accurate than e.g. IMERG, with the obvious downside of limited geographic coverage. MRMS natively has a resolution of $\ang{0.01}$. For computational reasons we subsample to \twentiethdegree for evaluation.

Rain gauges are yet another independent data source. We obtain quality controlled rain gauge data from Synoptic's derived precipitation service\footnote{https://synopticdata.com/}. Rain gauges provide reliable observations, yet suffer, just like weather stations, from an unequal global distribution (\cref{fig:fig_a1_station_distributions}). For all evaluations against rain gauges the predictions were interpolated bilinearly to the station locations.

\subsection{Baselines}

\paragraph{WeatherNext 2.}
For analysis evaluation, our key baseline is WeatherNext 2 \citep[WN2,][]{alet2025skillful} which represents the previous state-of-the-art in probabilistic medium-range forecasting. For the 2024 evaluations, we use a model version that has been trained until the end of 2023. For the real-time evaluation we use the publicly available WeatherNext 2 forecasts, which are based on a model trained until the end of 2025. 

\paragraph{ECMWF ENS.}
ENS has long been the gold standard in physics-based, ensemble medium-range prediction. Here we use ENS primarily as a baseline for variables not available in WN2. ENS has a resolution of \tenthdegree. We downloaded different subsets of ENS forecasts from different sources, and have noticed some discrepancies in performance depending on the source (see \cref{sec:app_ens} for details). This slightly penalizes ENS in some evaluations. However, since the gap between ENS and the AI-based models is generally quite large, we do not expect this to affect the key conclusions of our evaluations. For the real-time evaluations we use data from the ECMWF open-bucket available at \quarterdegree. This is clearly denoted in the legends and captions.

\paragraph{ECMWF AIFS ENS v2.}
AIFS ENS \citep{lang2026aifs} is another strong global AI-weather model. It has been upgraded to v2 on May 12, along with the switch to the 50r1 cycle~\citep{ecmwf2026aifsv2}. AIFS ENS consists of 50 members initialized from the perturbed ENS initial conditions, plus a control member initialized from the unperturbed initial conditions. For our real-time evaluation we use data from the ECMWF open-bucket that has been regridded from AIFS ENS's native N320 reduced Gaussian grid to \quarterdegree. Note that an hourly interpolation model was recently released for AIFS \citep{ingstad2026hourglass}. However, these forecasts are not yet readily available, so we were not able to evaluate them.

\subsection{Metrics}
\label{sec:metrics}

We use the continuous ranked probability score \citep[CRPS,][]{ferro2008effect} as our main evaluation metric. Lower CRPS values indicate better forecasts. See \cref{sec:app_eval:crps} for details.

\paragraph{Precipitation evaluation.}
Precipitation evaluation is a complex topic without a single agreed-upon best score to determine the "best" precipitation forecasts. CRPS serves as our lead metric, similar to ECMWF's headline metrics \citep{ecmwf2025eval}. To measure probabilistic skill at different precipitation rates, we additionally use the Brier Score \citep{glenn1950verification}. Note that we cap our evaluations at moderate accumulations of 4 mm/6h. For more extreme values, true positives become very sparse, making reliable evaluation difficult. For this reason, we leave the evaluation of extreme precipitation for future work.
We also compute reliability diagrams \citep{murphy1977reliability}, which measure how well predicted probabilities correspond, on average, with the actual occurrence frequency of events. Well calibrated forecasts should lie on the 1:1 line. Forecasts below the diagonal represent underdispersive, overconfident forecasts.
Finally, to assess precipitation skill for larger spatial regions, we use a pooled version of CRPS, where we either compute the score for the average or maximum values in a given region. This is similar to the pooled evaluations in \cite{price2025probabilistic} and \cite{alet2025skillful} but the pooling method differs slightly (\cref{app:pooled_crps}).

\paragraph{Latency-adjusted evaluation.}
WN3 produces a new forecast every hour, increasing update cadence and reducing operational forecast age compared to conventional models that run on 6-hour cycles. To quantify the skill improvement enabled by this higher frequency, we evaluate precipitation performance against an identical model constrained to 00/06/12/18 UTC initializations (denoted "6-hourly inits" in \cref{fig:figure_4_precip}c). Here, rather than looking at lead times relative to nominal initialization time, we look at operational lead times, where $t=0$ denotes the time at which the forecast is operationally available. We assume a latency of 7 hours (6 hours for raw input availability plus $\leq$1 hour for model execution and dissemination) for WN3. 

For example, consider an operational lead time of 2 hours at a decision time of 16 UTC (targeting valid time 18 UTC). The most recent WN3 forecast has a nominal initialization at 09 UTC (which takes as input HRES-fc3 initialized at 06 UTC, as well as geostationary satellite mosaics at 09 through 14 UTC), so the operational lead time of 2 hours corresponds to the model's nominal lead time of 9 hours. On the other hand, assuming 6-hourly initializations, the most recent forecast has a nominal initialization at 06 UTC, requiring a nominal lead time of 12 hours to target valid time 18 UTC. In this example, therefore, the hourly-initialized WN3 has a 3 hour advantage.

\paragraph{Cyclones evaluation.} 
For our cyclones evaluation we evaluate against the IBTrACS database as ground truth \citep{knapp2010ibtracs,knapp2024ibtracs}. WN3 and WN2 are both global models, so our evaluation includes cyclones from all basins. Following the evaluation protocol of \cite{alet2026operational} and the National Hurricane Center verification report \citep{cangialosi2026national}, we: (i) perform a homogeneous evaluation, whereby we only include a datapoint if both WN3 and WN2 predict it, and it also exists in the ground truth; (ii) include only datapoints where the cyclone nature was classified as tropical or subtropical, excluding disturbances, extratropical and unclassified systems. In addition, we account for the 6-hour lateness of the models' forecasts by performing the same `earlification' process from \cite{alet2026operational}, taking the forecast initialized from the previous 6-hour cycle, and applying a correction term to it based on operationally available TCVitals data \citep{tcvitals_format}.~\looseness=-1

\begin{figure}[ht]
    \centering\includegraphics[width=1.0\linewidth]{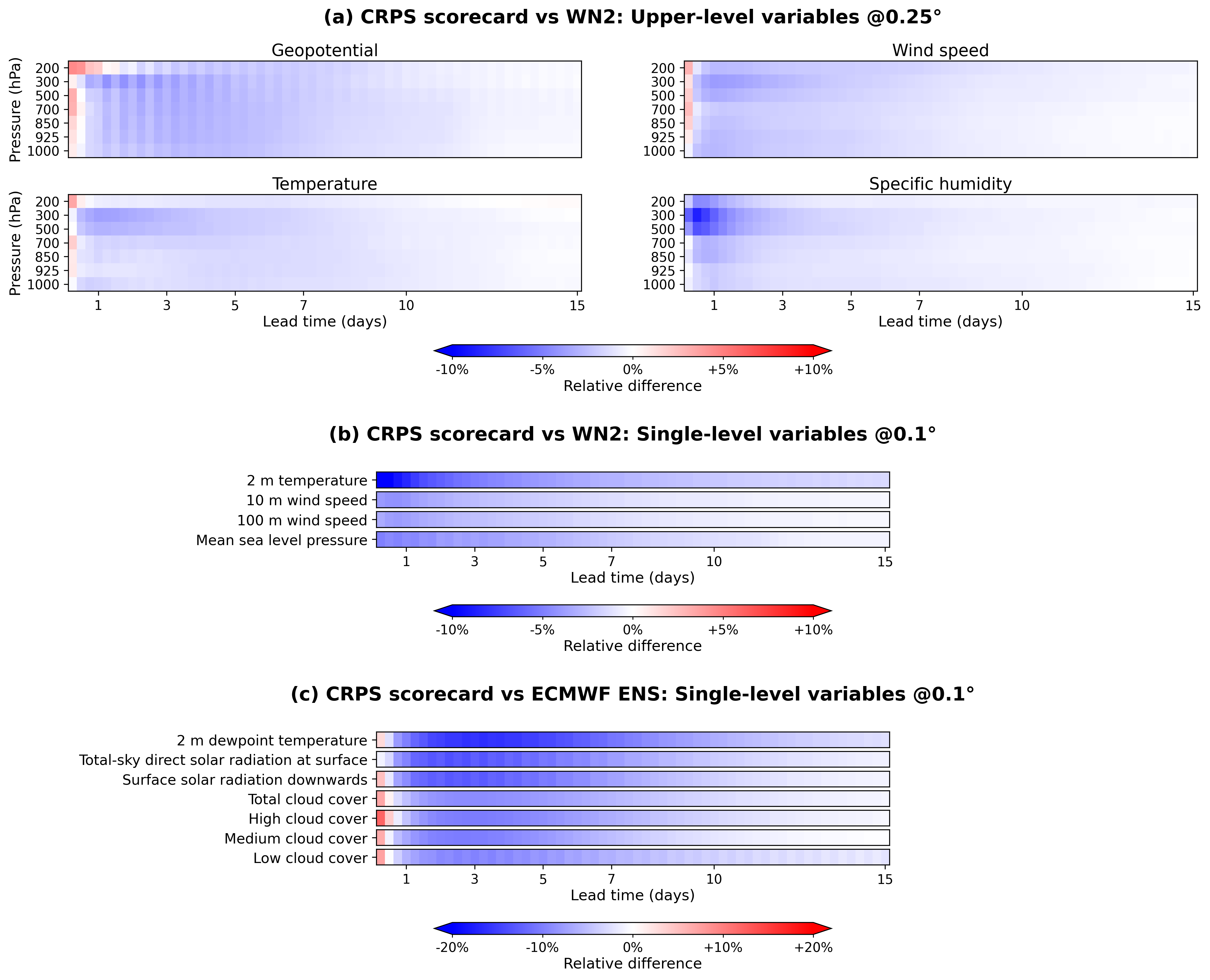}
    \caption{\textbf{Analysis results.} (a) Upper-level CRPS scorecard comparing WN3 to the previous state-of-the-art model WN2 at \quarterdegree. Blue squares indicate improvement. (b) Single-level variable CRPS scorecard of WN2 vs WN3 evaluated at \tenthdegree. WN2 forecasts were interpolated bilinearly. (c) Scorecard of variables not in WN2 against ECMWF ENS at \tenthdegree. The ground truth for (a), (b) and (c) is \hresfczero. All scores computed for 00/12 UTC initializations in 2024.
    }
    \label{fig:figure_2_analysis}
\end{figure}

\section{Results}
WN3 shows state-of-the-art performance across a variety of ground truths. In this section, we first focus on evaluations spanning a full year, 2024, starting with analysis scorecards, followed by evaluations against observational targets: weather stations, precipitation and tropical cyclones. Then, to evaluate the skill of our operationally running model version against baselines such as AIFS ENS, we show evaluations for a 6 week period in 2026. Finally, we discuss evaluations of the spatial-joint forecast distribution and artifacts in the model output.

\begin{figure}[t!]
    \centering\includegraphics[width=1.0\linewidth]{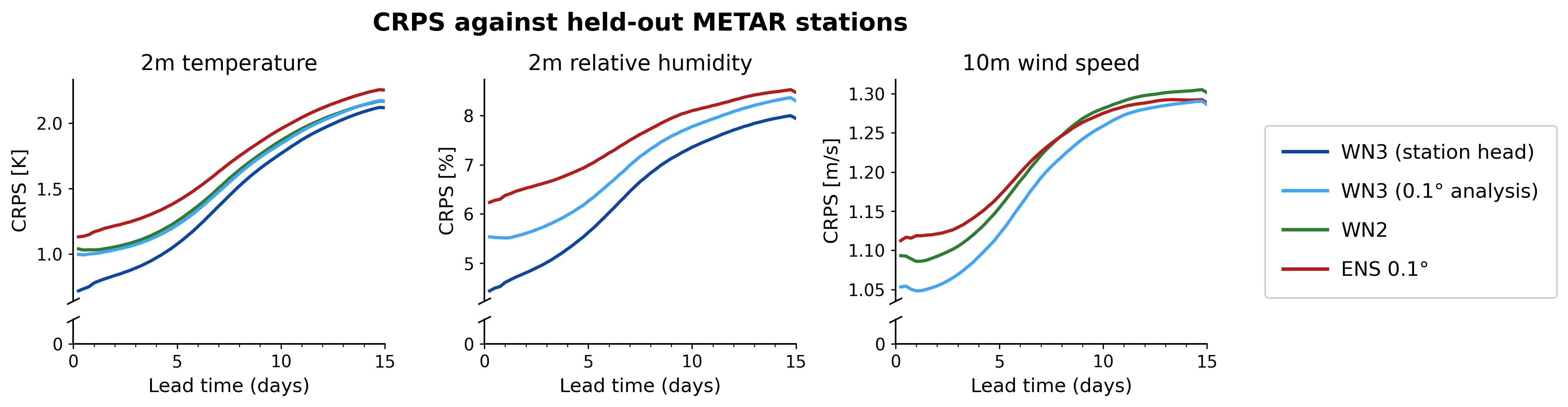}
    \caption{\textbf{Station evaluation results.} CRPS of 2m temperature, 2m relative humidity and 10m wind speed computed against METAR weather stations that have not been used during training. WN3 station head forecasts were queried at \twentiethdegree. All forecasts were bilinearly interpolated to the station locations. \quarterdegree and \tenthdegree 2m temperature forecasts are lapse-rate adjusted (see \cref{sec:gt}). All scores computed for 00/12 UTC initializations in 2024. For visual clarity, lines were smoothed with a 24 hour running average.
    }
    \label{fig:figure_3_stations}
\end{figure}

\subsection{Analysis evaluation}
WN3 outperforms the previous state-of-the-art global weather model WN2 on a majority of upper-level targets (\cref{fig:figure_2_analysis}a). Particularly large improvements can be seen for mid-to-upper-level humidity which hints at the additional information provided by the geostationary satellite imagery. For single-level variables, we evaluate against \tenthdegree data, which WN3 directly predicts, while we bilinearly interpolate the \quarterdegree WN2 forecasts. WN3 outperforms WN2 across all surface variables, with the largest improvements seen for 2m temperature (\cref{fig:figure_2_analysis}b). WN3 still outperforms WN2 when only evaluated on common ($\ang{0.5}$) grid points, albeit by a smaller margin (\cref{fig:fig_appendix_analysis_relative_lines_2024}). This shows that the improvement comes from both better general model skill and increased resolution. Comparisons against ENS show even larger improvements (\cref{fig:fig_appendix_analysis_relative_lines_2024}). For variables that were not available in WN2, we use ENS as our primary baseline (\cref{fig:figure_2_analysis}c). Here as well, WN3 shows substantial improvements in CRPS skill across all variables. Note that for cumulative variables (ssrd and fdir) we are using the 6h WN3 variables for this evaluation (\cref{sec:app_model_formulation}).

The only noteworthy exceptions to WN3's superior performance on analysis forecasting are the 6h (and occasionally 12h) lead times for a subset of variables. Further research and ablations are needed to definitively understand the causes of this phenomenon. However, we note that due to the roughly 7h latency in global forecast systems, nominal 6-hour forecasts are never actually used operationally, and so 6h performance has little effect in practice.

The medium-range improvements of roughly 5\% on upper-level variables with respect to WN2 approximately translate to 6 hours of additional forecast lead time at the same level of skill. Traditionally, weather forecast skill has progressed by one day of forecast skill per decade \citep{bauer2015quiet}. This improvement over WN2, published a bit more than a year ago, demonstrates the continued accelerated pace of progress enabled by AI-weather models. However, one of WN3's major contributions is to go beyond a pure focus on analysis-based medium-range skill towards observations.

\subsection{Weather station evaluation}
Evaluation against global METAR weather station data shows that WN3's station head predictions have substantially smaller error compared to analysis-based forecasts (\cref{fig:figure_3_stations}, and \cref{fig:fig_appendix_hourly_station_evals_48h} for evaluation of hourly lead times $\leq 2$ days). For short lead times, the station head improves CRPS for 2m temperature by up to 30\% compared to WN2 and 40\% compared to ENS. Similar improvements with respect to ENS hold for 2m relative humidity (WN2 does not predict 2m dewpoint temperature). All evaluations shown here are based on stations not seen during training. The station head does overfit to training stations but only by a few percent, indicating strong spatial generalization (\cref{fig:fig_appendix_station_holdout_abs_crps}). 

The WN3 \tenthdegree analysis predictions also show strong performance, demonstrating the benefit of increased resolution. For 10m wind speed, WN3 leads to a 5\% reduction in CRPS at early lead times. Note that because of the noise in the wind speed observations, the base error is quite high compared to the increase in the forecast error with lead time. 

As mentioned in \cref{sec:gt}, we apply a lapse rate-adjustment to the \quarterdegree and \tenthdegree 2m temperature predictions. This improves the coarse forecasts the most. For our \twentiethdegree station head predictions, however, this additional post-processing turned out to be unnecessary, which highlights the ability of WN3 to combine traditionally separate stages of the weather forecasting pipeline into a single model.

\subsection{Precipitation evaluation}
\label{sec:precip_eval}

\begin{figure}[t!]
    \centering\includegraphics[width=1.0\linewidth]{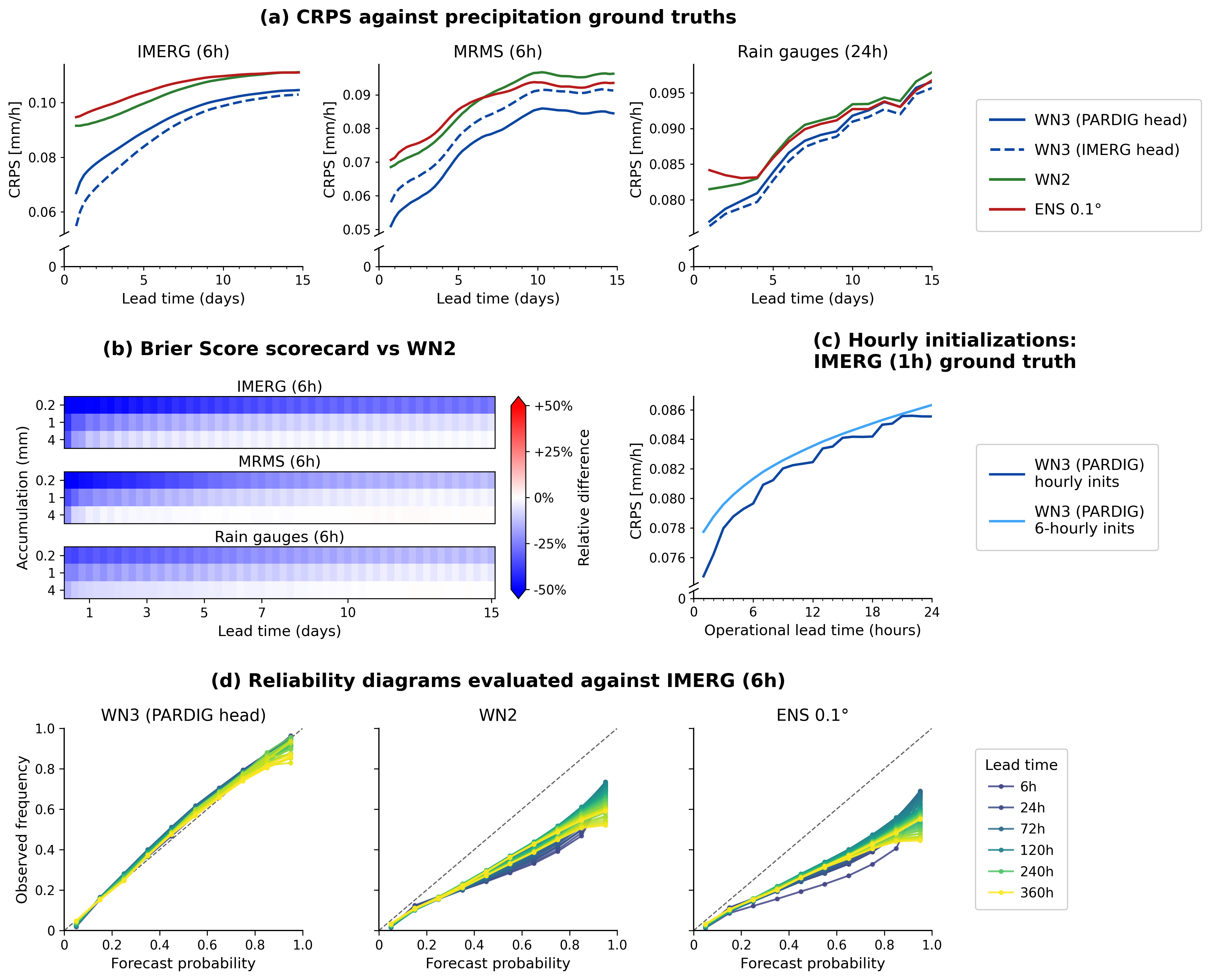}
    \caption{\textbf{Precipitation results.} (a) CRPS of precipitation forecasts evaluated against IMERG (6h accumulation), MRMS (6h) and rain gauges (24h). Computed for 00/12 UTC initializations. (b) Brier score scorecard for the WN3 PARDIG head with respect to WN2 for different thresholds evaluated against IMERG, MRMS and rain gauges for 6h accumulation. Computed for 00/06/12/18 UTC initializations. (c) Latency-adjusted CRPS comparing hourly with 6-hourly initializations evaluated against IMERG for 1h accumulations. x-axis shows operational lead time (see \cref{sec:metrics}). (d) Reliability diagrams of the WN3 PARDIG head, WN2 and ENS for 1mm / 6h accumulation evaluated against IMERG. Reliability diagrams computed for 00/12 UTC initializations.
    }
    \label{fig:figure_4_precip}
\end{figure}

WN3's PARDIG head predictions greatly improve precipitation skill when evaluated against a range of ground truths. Evaluations against global probabilistic baselines WN2 and ENS show a reduction in CRPS by up to 60\% for IMERG, 30\% for MRMS and 10\% for rain gauge measurements for early lead times (\cref{fig:figure_4_precip}a). The IMERG head naturally shows lower errors when using IMERG as a ground truth, but performs worse against MRMS, and similarly against 24h rain gauge measurements.
More fine-grained analysis using the Brier Score for different precipitation rates in \cref{fig:figure_4_precip}b shows that improvements are most pronounced for low precipitation rates around the rain/no-rain threshold; \cref{fig:fig_appendix_precip_brier_scores} shows comparisons that include ENS and both WN3 precipitation heads. For higher rates absolute Brier Score values become very small and less reliable at longer lead times, highlighting the need for more specific precipitation metrics.

For precipitation forecasts, it is especially important that forecasts are well calibrated, meaning that predicted probabilities of rain, on average, match the actual occurrence frequency. Reliability diagrams for the WN3 PARDIG head, as well as WN2 and ENS, show that WN3 is better calibrated against IMERG compared to previous global models, which are overconfident in their predictions (\cref{fig:figure_4_precip}d). Notably, calibration also remains stable throughout the 15 day forecast. The pronounced underdispersion of ENS at 6 hours lead time is likely a symptom of model spin-up. 

Finally, we focus on the impact of hourly initializations enabled by using low-latency geostationary satellite observations (\cref{fig:figure_4_precip}c and \cref{fig:fig_appendix_precip_rapid_refresh_init_time}). For this we compare the latency adjusted skill of the 6-hourly WN3 PARDIG head predictions to the hourly updates. The hourly forecasts show improved skill, particularly for early lead times. The lead time gained is 2--3 hours, which is in line with expectations since the average latency advantage of hourly over 6-hourly initializations is roughly 3 hours. This gain is important for quickly developing storms.

\begin{figure}[t!]
    \centering\includegraphics[width=1.0\linewidth]{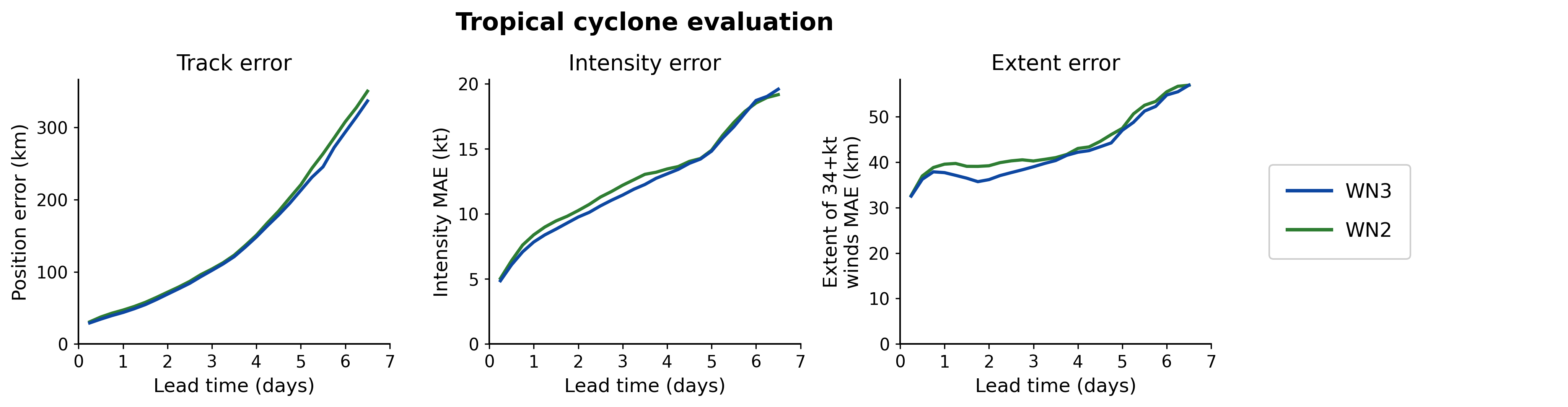}
    \caption{\textbf{Cyclone results.} Ensemble mean track, intensity and extent error for WN2 and WN3. Intensity is defined as the maximum 1-minute sustained wind speed. Extent error is the mean absolute error of 34kt wind radii forecasts averaged over all quadrants. Evaluations for 00/06/12/18 UTC initializations in 2024.
    Forecasts are homogenized across WN2 and WN3, and include cyclones across all basins across the globe.
    We also restrict to data points classified as tropical or subtropical in nature in IBTrACS \citep{knapp2010ibtracs,knapp2024ibtracs}, excluding extratropical or unclassified systems.
    }
    \label{fig:figure_5_cyclones}
\end{figure}

\subsection{Tropical cyclone evaluation} 

We follow the evaluation protocol of~\citet{alet2026operational} and compare WN3 against WN2 in a global homogeneous evaluation.
WN2 is a modified version of WeatherNext Cyclones (WN-C) from~\citet{alet2026operational} with the same training protocol and nearly identical training data, except it also uses 100-metre wind as inputs and targets.
Apart from this difference, the performance of WN2 and WN-C are extremely similar on both atmospheric and cyclone variables, and were the state-of-the-art in AI-based cyclone forecasting in the 2025 season.
\cref{fig:figure_5_cyclones} shows relatively small, but consistent improvements in mean absolute error of both the ensemble mean track and intensity forecasts.
Extent improvements are more sizeable between 1- and 3-day lead times.
At the same time, we also observed that WN3 tended to be generally more under-spread than WN2 on both intensity and extent (\cref{fig:fig_appendix_cyclone_spread_skill,fig:fig_appendix_cyclone_extent_eval}). This under-spreading may be due to the larger size of the model which could induce some amount of overfitting on the cyclone variables.

\begin{figure}[!ht]
    \centering\includegraphics[width=1.0\linewidth]{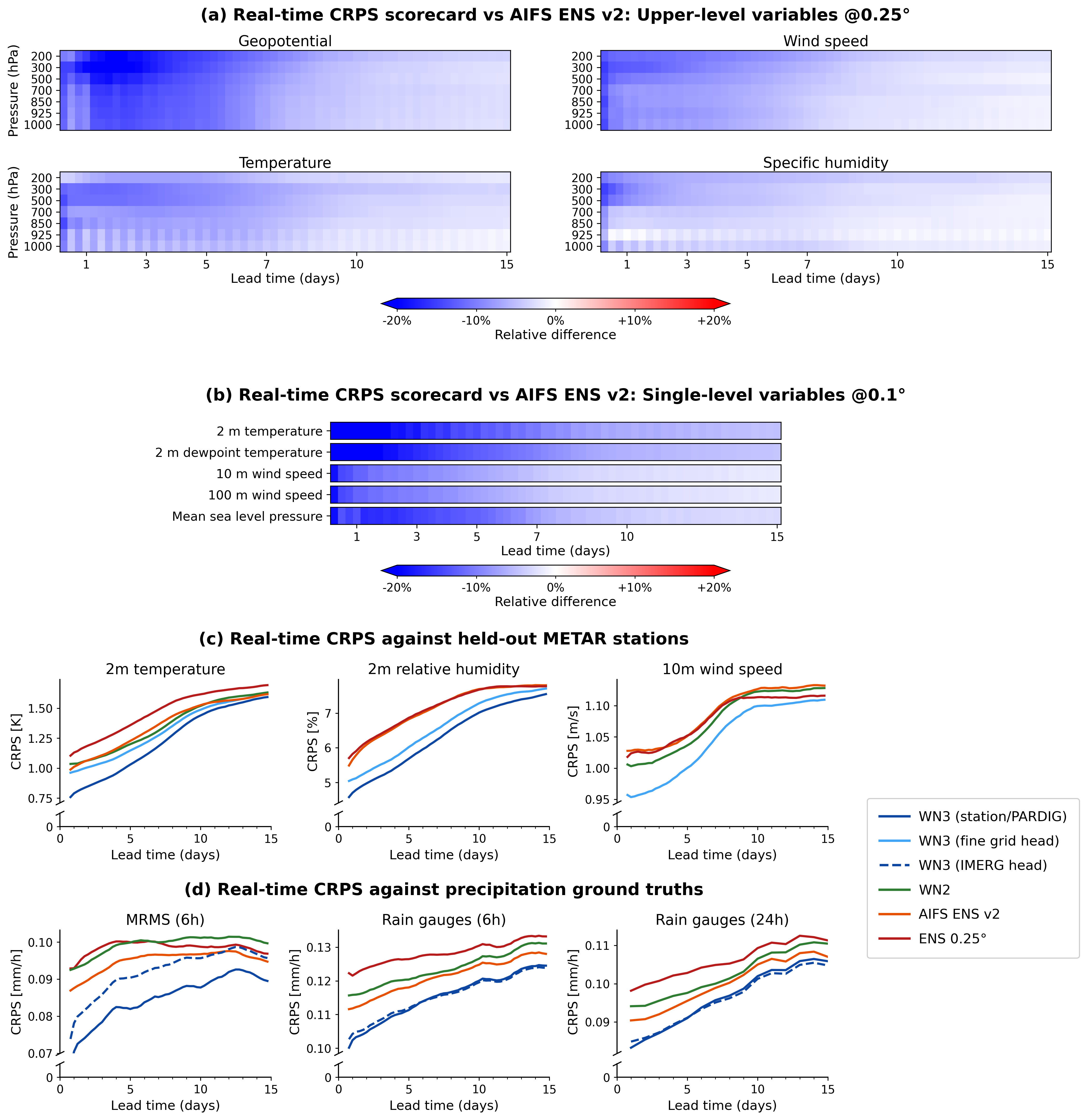}
    \caption{\textbf{Real-time evaluation} (a) Upper level CRPS scorecard with respect to AIFS ENS v2 at \quarterdegree. Ground truth for AIFS ENS is the AIFS ENS Control fc0, while the ground truth for WN3 is \hresfczero; (b) Single-level evaluations at \tenthdegree. \quarterdegree AIFS forecasts are bilinearly interpolated. (c) Evaluations against METAR surface stations not seen during WN3 training. \quarterdegree and \tenthdegree 2m temperature forecasts are lapse-rate adjusted. (d) Precipitation evaluations against MRMS (6h accumulation) and rain gauges (6 and 24h). For visual clarity, 6h lines in (c) and (d) have been smoothed with a 24 hour running average.  All evaluations are for July 1 to August 11 on 00/12 UTC initializations.
    }
    \label{fig:figure_6_aifs}
\end{figure}

\subsection{Real-time evaluation}
To fairly compare the actual operational skill of the leading AI weather models, we run a quasi-real time evaluation using the operational data feeds of WN3, WN2 and AIFS ENS for the 6-week period July 1 to Aug 11 2026. This time period was chosen to allow us to evaluate our production model, which was trained until June 30 2026 in order to include 50r1 data. While 6 weeks make for a relatively small sample size, clear trends are still visible but it is important to not over-interpret smaller signals. WN2's operational checkpoint was trained until the end of 2024, and thus does not include 50r1, while both WN3 and AIFS ENS v2 include some of it in their training data.

WN3 outperforms AIFS ENS across all upper level variables with an average improvement of roughly 10\% in the first forecast week (\cref{fig:figure_6_aifs}a). For single-level variables, we evaluate against \tenthdegree \hresfczero ground truth, interpolating the \quarterdegree AIFS ENS forecasts. There, WN3 has a substantial advantage. Part of this is due to differences in interpolation and the choice of ground truth (see \cref{sec:app_data} and \cref{fig:fig_appendix_analysis_relative_lines_real_time}). Even when accounting for these differences, however, WN3 outperforms AIFS ENS on the majority of variables and lead times. On station evaluations the WN3 station head shows a substantial reduction in 2m temperature and relative humidity error with respect to AIFS ENS (\cref{fig:figure_6_aifs}c). The WN3 \tenthdegree analysis predictions also perform strongly when evaluated against stations with a 5\% improvement in wind speed CRPS over the baselines. AIFS ENS is roughly on par with WN2 for this period on 2m temperature and slightly worse on 10m wind speed.

\begin{figure}[th!]
    \centering\includegraphics[width=1.0\linewidth]{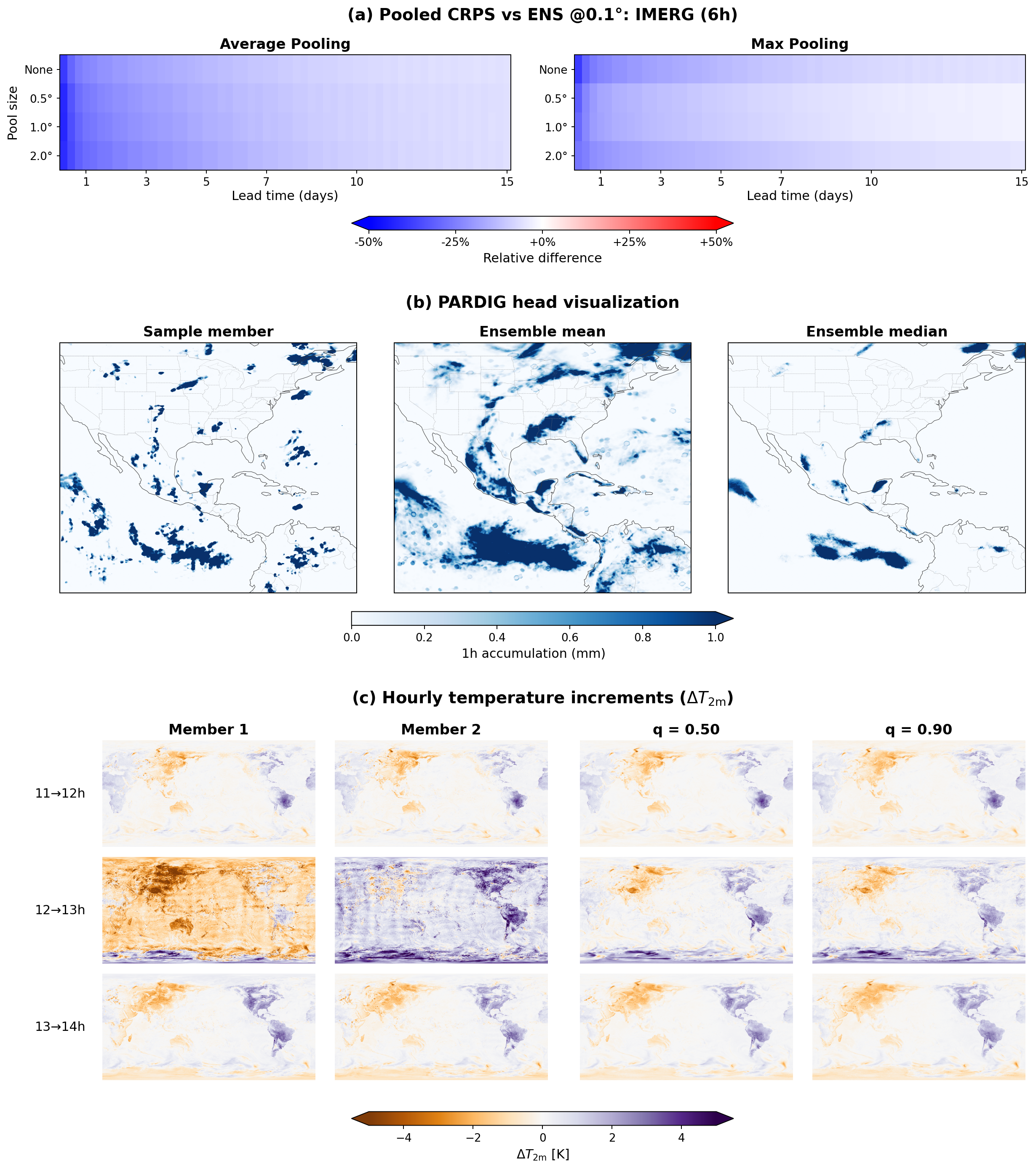}
    \caption{\textbf{Pooled metrics and artifact visualization.} (a) 6h precipitation accumulation CRPS scorecard of WN3 PARDIG head with respect to ENS evaluated against IMERG at different pooling sizes for average and max pooling. (b) Example of artifacts in 24h PARDIG head forecasts (1h accumulation) for a single member, the 64-member ensemble mean and median for 24h forecast lead time (note that the color bar saturates at 1mm for the purposes of visualization). (c) Visualization of hourly temporal differences in the WN3 station head for 2m temperature. 12 to 13h exemplifies the transition between outer 6h time steps; (top two rows) 2 individual ensemble members; (bottom three rows) Temporal differences in ensemble quantiles ($q=$ 0.5 and 0.9). All examples for 2026-07-01T00.
    }
    \label{fig:figure_7_artifacts}
\end{figure}

For precipitation, we evaluate against MRMS (6h) and rain gauges (6 and 24h). IMERG Final was not available for this real-time period. Because of the limited sample size, the precipitation evaluations are relatively noisy, especially for MRMS which only has limited geographic coverage. For MRMS, the PARDIG head has the lowest CRPS compared to the IMERG head and other baselines by some margin. For the rain gauge evaluations the PARDIG and IMERG heads are roughly on par, with a substantial gap to AIFS ENS, which itself performs strongly against WN2 and ENS.

\subsection{Joint forecast skill and artifacts}

Despite being trained to optimize marginal skill, FGN is able to produce largely realistic joint structures \citep{alet2025skillful}. This is evident in its ability to produce reliable forecasts of cyclones \citep{alet2026operational}, derived (wind speed and relative humidity) and accumulated variables (6 or 24h precipitation) as well as realistic-looking forecast visuals (\cref{fig:fig_appendix_fine_grid_analysis_global}). Further, spatial metrics show strong performance across spatial scales (see Fig. 3 in  \citealp{alet2025skillful}). Here, we perform a spatially pooled evaluation of the WN3 PARDIG head precipitation forecasts (\cref{fig:figure_7_artifacts}a) for 6h accumulations. No systematic degradation of forecast skill is evident for larger pooling sizes. This is important for making reliable probabilistic predictions of cumulative precipitation over larger areas, used for example in flood forecasting.

Nevertheless, spatial and temporal artifacts are visible in individual WN3 samples. These issues arise because optimizing for marginal CRPS allows the model to `cheat' on the covariance structure to maximize point-wise skill. One manifestation of this behaviour, first noted in WN2 \citep[Fig.~5 of ][]{alet2025skillful}, is the appearance of distinct hexagonal patterns that reflect the model's underlying grid mesh structure. While visible to some degree across most variables, they are most severe in precipitation and station outputs. \cref{fig:figure_7_artifacts}b illustrates these patterns in the PARDIG head, where we have chosen a colormap with a deliberately low maximum hourly accumulation of 1mm in order to show the artifacts clearly (see \cref{fig:fig_appendix_station_head_visualizations} for station head examples). These artifacts are still visible in the ensemble mean, but less so in the ensemble median. We find that the PARDIG head has more pronounced artifacts than the IMERG head and believe this is due to PARDIG having a sharper distribution, leading to higher overall skill for the PARDIG head but making individual member artifacts more evident (\cref{fig:fig_appendix_precip_heads_comparison}).

Further, temporal discontinuities across the 6h outer time step boundaries are evident in the station output head, and to some degree the PARDIG and IMERG output heads. (Note that this does not apply to the hourly analysis output, which has no noticeable discontinuities.) The station output head also exhibits a per-ensemble-member global bias, where one sample will tend to be warmer or colder globally compared to the ensemble mean. This global bias changes with each 6 hour time step as a new noise vector is drawn. \cref{fig:figure_7_artifacts}c shows the hourly temporal difference in 2m temperature station head predictions $\Delta T_{2m}$, which clearly shows the large jumps across 6h boundaries (12h to 13h). However, marginal statistics, such as quantiles, are largely artifact-free (except in Antarctica which is badly constrained in the station head due to the lack of observation data).  Since many downstream applications rely primarily on ensemble statistics like median or threshold-exceedance probabilities, these forecasts remain highly usable.

\section{Discussion}
WeatherNext 3 offers substantial improvements over previous operational global weather models in forecast skill and capabilities. In terms of raw medium-range forecast skill, it sets a new state-of-the-art for probabilistic weather prediction, outperforming both its predecessor WN2 and ECMWF's AIFS ENS v2 on a vast majority of analysis metrics. WN3 also produces output of single-level variables at \tenthdegree, matching the resolution of the best traditional global NWP system, ECMWF ENS. In addition, leveraging low-latency satellite input allows WN3 to be initialized every hour, rather than every 6 hours as is typical for most global weather models, allowing the model to adapt to rapidly changing weather phenomena, which delivers a boost in precipitation skill in early lead times.

Another jump in performance comes from directly training against high-quality observation data, circumventing known biases and inaccuracies in analysis data. For precipitation, we use our own precipitation reanalysis, PARDIG, derived from global space-borne radar data. This results in improved and more reliable precipitation forecasts, evaluated against a range of ground truths. For on-the-ground temperature and humidity, weather stations serve as the gold standard observations. Traditionally, sparse data sources have been hard to use in machine learning models. WN3's architecture allows us to train directly against sparse data, yet still produce predictions anywhere on the globe provided easily available metadata, such as elevation and a land/sea mask. These predictions show a substantial reduction in error, even when evaluated against unseen locations. It is important to highlight that machine learning-based post-processing approaches for correcting global model output to station observations have been an established part of the weather forecasting toolkit for many decades \citep{glahn1972use, gneiting2005calibrated, hewson2021low}. By directly incorporating station calibration into the model, WN3 has the advantage of not needing a separate calibration stage. This reduces processing time and allows the station head to directly tap into the core model latent. Another feature of WN3's station head is its strong spatial generalization. Post-processing methods that generalize beyond the training stations are an active area of research but, again, require training, fine-tuning and running a separate model \citep{vannitsem2021statistical}.

We believe that WN3 represents a major step forward for AI-based weather predictions by going beyond relying purely on analysis and utilizing information-dense, low-latency observation data. Physics-based analyses provide an uninterrupted, dense dataset perfect for training AI models. The existence of ERA5, for instance, has been a major contributor to the rapid advancement of AI weather models. However, it has also long been acknowledged that analysis only represents a best guess of the atmosphere, unable to use all observation sources to their full extent and exhibiting biases, especially for small scale processes. Here we demonstrate that augmenting an analysis-based model with observations in input and output allows us to alleviate these shortcomings and combine the best of both worlds. 

\FloatBarrier

\section*{Acknowledgments}

In alphabetical order, we thank Abby Bullock, Alex Wilkins, Aliyah Bond, Avinatan Hassidim, Bradley Goldstein, Christian Wagner, Daniel Rothenberg, Daniel Worrall, Darshan Prajapati, David Landry, Devaja Shah, Drew Bollinger, Drew Purves, Elinor Kruse, Emily Morris, Emma Yousif, Gaby Pearl, Hans Mohrmann, Jash Rana, Jenny Shepard, Jessica Sapick, Juanita Bawagan, Julian Green, Kasia Mohammed, Kunal Shah, Leen Verburgh, Marc Deisenroth, Marcus Trail, Natalie Williams, Piyush Ingale, Rachel Stigler, Rahul Mahrsee, Raia Hadsell, Reuven Sayag, Ryan Keisler, Shail Parekh, Shubham Joshi, Shubham Kumar, Stephanie Sanchez, Thomas Turnbull, Zach Hynes, and Zoubin Ghahramani.

We also thank ECMWF and NOAA for providing invaluable datasets to the research community, and the experts at NHC, UKMet and WNI.

\section*{Disclaimer}
\textbf{Experimental Forecasts:} WeatherNext is an automated, experimental AI system under active research. Predictions are provided "as-is" for informational purposes and are not official severe weather warnings. For protection of life and property, always defer to official alerts from your local emergency authorities and national meteorological services.  Use of WeatherNext and its output are subject to our \href{https://storage.googleapis.com/weathernext-public/terms-of-use.pdf}{Terms of Service}.

This document is based on data and products of the European Centre for Medium-Range Weather Forecasts (ECMWF).

\newpage

\bibliography{main}

@article{ecmwf2026aifsv2,
  author  = {{ECMWF}},
  title   = {Implementation of {AIFS} v2},
  journal = {ECMWF Newsletter},
  volume  = {187},
  pages   = {16--22},
  year    = {2026},
  url     = {https://www.ecmwf.int/en/newsletter/187/news/implementation-aifs-v2}
}

@article{diaconu2026otter,
  title={Otter Weather: Skillful and Computationally Efficient Medium-Range Weather Forecasting},
  author={Diaconu, Cristiana and Scholz, Jonas and Shysheya, Aliaksandra and Markou, Stratis and Mukhopadhyay, Payel and Cranmer, Miles and Turner, Richard E},
  journal={arXiv preprint arXiv:2606.26421},
  year={2026}
}

@article{cachay2026u,
  title={U-Cast: A Surprisingly Simple and Efficient Frontier Probabilistic AI Weather Forecaster},
  author={Cachay, Salva R{\"u}hling and Watson-Parris, Duncan and Yu, Rose},
  journal={arXiv preprint arXiv:2604.09041},
  year={2026}
}

@techreport{cangialosi2026national,
  author      = {Cangialosi, John P. and Martinez, Jon},
  title       = {National Hurricane Center Forecast Verification Report: 2025 Hurricane Season},
  institution = {National Hurricane Center (NHC)},
  year        = {2026},
  url         = {https://www.nhc.noaa.gov/verification/pdfs/Verification_2025.pdf}
}

@article{kochkov2024neural,
  title={Neural general circulation models for weather and climate},
  author={Kochkov, Dmitrii and Yuval, Janni and Langmore, Ian and Norgaard, Peter and Smith, Jamie and Mooers, Griffin and Kl{\"o}wer, Milan and Lottes, James and Rasp, Stephan and D{\"u}ben, Peter and others},
  journal={Nature},
  volume={632},
  number={8027},
  pages={1060--1066},
  year={2024},
  publisher={Nature Publishing Group UK London}
}

@online{windborne2026weathermesh6,
  author       = {{WindBorne Systems}},
  title        = {What's New in WeatherMesh-6},
  year         = {2026},
  month        = {June},
  day          = {1},
  url          = {https://windbornesystems.com/blog/introducing-wm-6},
  urldate      = {2026-09-02}
}

@techreport{danielson2011gmted2010,
  author      = {Danielson, Jeffrey J. and Gesch, Dean B.},
  title       = {Global Multi-resolution Terrain Elevation Data 2010 ({GMTED2010})},
  institution = {U.S. Geological Survey},
  series      = {Open-File Report},
  number      = {2011-1073},
  year        = {2011},
  pages       = {26},
  doi         = {10.3133/ofr20111073},
  url         = {https://doi.org/10.3133/ofr20111073}
}

@misc{ecmwf_hres_2023,
  author       = {{ECMWF}},
  title        = {{ECMWF High Resolution (HRES) Analysis Dataset}},
  year         = {2023},
  publisher    = {ECMWF},
  url          = {https://www.ecmwf.int},
  note         = {Available under Creative Commons Attribution 4.0 International (CC BY 4.0)}
}

@misc{zanaga2022worldcover2021,
  author       = {Zanaga, Daniele and
                  Van De Kerchove, Ruben and
                  Daems, Dirk and
                  De Keersmaecker, Wanda and
                  Brockmann, Carsten and
                  Kirches, Grit and
                  Wevers, Jan and
                  Cartus, Oliver and
                  Santoro, Maurizio and
                  Fritz, Steffen and
                  Lesiv, Myroslava and
                  Herold, Martin and
                  Tsendbazar, Nandin-Erdene and
                  Xu, Panpan and
                  Ramoino, Fabrizio and
                  Arino, Olivier},
  title        = {{ESA WorldCover 10 m 2021 v200}},
  year         = {2022},
  publisher    = {Zenodo},
  version      = {v200},
  doi          = {10.5281/zenodo.7254221},
  url          = {https://doi.org/10.5281/zenodo.7254221}
}

@article{ingstad2026hourglass,
  title={HourGlass: A probabilistic data-driven temporal downscaler for global and regional weather forecasting},
  author={Ingstad, Magnus Sikora and Clare, Mariana CA and Ersland, Olav and Gahlen, Vera and Haugen, H{\aa}vard Homleid and Miralles, Oph{\'e}lia and Nordhagen, Even M and Nipen, Thomas N and Seierstad, Ivar A and Bremnes, John Bj{\o}rnar and others},
  journal={arXiv preprint arXiv:2607.11457},
  year={2026}
}

@misc{ecmwf_dissemination_schedule,
  author       = {{ECMWF}},
  title        = {{Dissemination Schedule}},
  howpublished = {\url{https://confluence.ecmwf.int/spaces/DAC/pages/272310483/Dissemination+schedule}},
  year         = {2026},
  note         = {Accessed: 2026-09-02}
}

@misc{ecmwf2025eval,
  author = {Thomas Haiden and Martin Janousek and Frédéric Vitart and Fernando Prates and Michael Maier-Gerber and Cathy Wing Yi Li and Matthieu Chevallier},
  title = {Evaluation of ECMWF forecasts},
  year = {2025},
  journal = {Technical Memoranda},
  number = {931},
  month = {09/2025},
  publisher = {ECMWF},
  address = {Reading},
  doi = {10.21957/51e665e5c8},
}

@online{tcvitals_format,
  author = {{NCEP}},
  title = {{NCEP, cited 2011: Format of the tropical cyclone vital statistics records}},
  url = {http://www.emc.ncep.noaa.gov/mmb/data_processing/tcvitals_description.htm},
  urldate = {2025-12-20},
  year={2025},
}

@misc{knapp2024ibtracs,
    author = {Gahtan, J. and Knapp, K. R. and Schreck, C. J. and Diamond,  H. J. and Kossin, J. P. and Kruk, M. C.},
    year = {2024}, 
    title = {International Best Track Archive for Climate Stewardship (IBTrACS) Project, Version 4r01}, 
    howpublished= {NOAA National Centers for Environmental Information. doi:10.25921/82ty-9e16},
}

@article{hersbach2020era5,
  author    = {Hersbach, Hans and Bell, Bill and Berrisford, Paul and Hirahara, Shoji and Hor{\'a}nyi, Andr{\'a}s and Mu{\~n}oz-Sabater, Joaqu{\'i}n and Nicolas, Julien and Peubey, Carole and Radu, Raluca and Schepers, Dinand and Simmons, Adrian and Soci, Cornel and Abdalla, Saleh and Abellan, Xavier and Balsamo, Gianpaolo and Bechtold, Peter and Biavati, Gionata and Bidlot, Jean and Bonavita, Massimo and De Chiara, Giovanna and Dahlgren, Per and Dee, Dick and Diamantakis, Michail and Dragani, Rossana and Flemming, Johannes and Forbes, Richard and Fuentes, Manuel and Geer, Alan and Haimberger, Leopold and Healy, Sean and Hogan, Robin J. and H{\'o}lm, El{\'i}as and Janiskov{\'a}, Marta and Keeley, Sarah and Laloyaux, Patrick and Lopez, Philippe and Lupu, Cristina and Radnoti, Gabor and de Rosnay, Patricia and Rozum, Iryna and Vamborg, Freja and Villaume, Sebastien and Th{\'e}paut, Jean-No{\"e}l},
  title     = {The {ERA5} Global Reanalysis},
  journal   = {Quarterly Journal of the Royal Meteorological Society},
  volume    = {146},
  number    = {730},
  pages     = {1999--2049},
  year      = {2020},
  doi       = {10.1002/qj.3803}
}

@article{freeman2017icoads,
  author    = {Freeman, Eric and Woodruff, Scott D. and Worley, Steven J. and Lubker, Sandra J. and Kent, Elizabeth C. and Angel, William E. and Berry, David I. and Flick, Philip and Freeman, Reynolds and Hutchins, Peter G. and Kolper, Ronald and Reynolds, Richard W. and Smith, Shawn R.},
  title     = {{ICOADS Release 3.0}: A Major Update to the Historical Marine Climate Record},
  journal   = {International Journal of Climatology},
  volume    = {37},
  number    = {5},
  pages     = {2211--2232},
  year      = {2017},
  doi       = {10.1002/joc.4775}
}

@misc{icoads_release3_dataset,
  author       = {{Research Data Archive at NCAR} and {NOAA NCEI}},
  title        = {{International Comprehensive Ocean-Atmosphere Data Set (ICOADS) Release 3, Individual Observations}},
  howpublished = {NSF National Center for Atmospheric Research and NOAA},
  year         = {2016},
  doi          = {10.5065/D6ZS2TR3}
}

@article{knapp2010ibtracs,
  author    = {Knapp, Kenneth R. and Kruk, Michael C. and Levinson, David H. and Diamond, Howard J. and Neumann, Charles J.},
  title     = {The {International Best Track Archive for Climate Stewardship (IBTrACS)}: Unifying Tropical Cyclone Data},
  journal   = {Bulletin of the American Meteorological Society},
  volume    = {91},
  number    = {3},
  pages     = {363--376},
  year      = {2010},
  doi       = {10.1175/2009BAMS2755.1}
}

@misc{noaa_nssl_mrms,
  author       = {{NOAA National Severe Storms Laboratory} and {NCEP}},
  title        = {{Multi-Radar Multi-Sensor (MRMS) System Quantitative Precipitation Estimation (QPE)}},
  howpublished = {NOAA NSSL and NCEP},
  year         = {2024},
  url          = {https://www.nssl.noaa.gov/projects/mrms/}
}

@article{huffman2023nasa,
  title={Nasa global precipitation measurement (GPM) integrated multi-satellite retrievals for GPM (IMERG) version 07},
  author={Huffman, George J and Bolvin, David T and Braithwaite, D and Hsu, K and Joyce, R and Kidd, C and Nelkin, E and Xie, P},
  journal={Algorithm theoretical basis document (ATBD) version},
  volume={47},
  year={2023}
}

@article{rasp2024weatherbench,
  title={WeatherBench 2: A benchmark for the next generation of data-driven global weather models},
  author={Rasp, Stephan and Hoyer, Stephan and Merose, Alexander and Langmore, Ian and Battaglia, Peter and Russell, Tyler and Sanchez-Gonzalez, Alvaro and Yang, Vivian and Carver, Rob and Agrawal, Shreya and others},
  journal={Journal of Advances in Modeling Earth Systems},
  volume={16},
  number={6},
  pages={e2023MS004019},
  year={2024},
  publisher={Wiley Online Library}
}

@article{alet2025skillful,
  title={Skillful joint probabilistic weather forecasting from marginals},
  author={Alet, Ferran and Price, Ilan and El-Kadi, Andrew and Masters, Dominic and Markou, Stratis and Andersson, Tom R and Stott, Jacklynn and Lam, Remi and Willson, Matthew and Sanchez-Gonzalez, Alvaro and others},
  journal={arXiv preprint arXiv:2506.10772},
  year={2025}
}

@article{price2025probabilistic,
  title={Probabilistic weather forecasting with machine learning},
  author={Price, Ilan and Sanchez-Gonzalez, Alvaro and Alet, Ferran and Andersson, Tom R and El-Kadi, Andrew and Masters, Dominic and Ewalds, Timo and Stott, Jacklynn and Mohamed, Shakir and Battaglia, Peter and others},
  journal={Nature},
  volume={637},
  number={8044},
  pages={84--90},
  year={2025},
  publisher={Nature Publishing Group UK London}
}

@article{vaughan2024aardvark,
  title={Aardvark weather: end-to-end data-driven weather forecasting},
  author={Vaughan, Anna and Markou, Stratis and Tebbutt, Will and Requeima, James and Bruinsma, Wessel P and Andersson, Tom R and Herzog, Michael and Lane, Nicholas D and Chantry, Matthew and Hosking, J Scott and others},
  journal={arXiv preprint arXiv:2404.00411},
  year={2024}
}

@article{soci2024era5,
  title={The ERA5 global reanalysis from 1940 to 2022},
  author={Soci, Cornel and Hersbach, Hans and Simmons, Adrian and Poli, Paul and Bell, Bill and Berrisford, Paul and Hor{\'a}nyi, Andr{\'a}s and Mu{\~n}oz-Sabater, Joaqu{\'\i}n and Nicolas, Julien and Radu, Raluca and others},
  journal={Quarterly Journal of the Royal Meteorological Society},
  volume={150},
  number={764},
  pages={4014--4048},
  year={2024},
  publisher={Wiley Online Library}
}

@article{lang2026aifs,
  title={AIFS-CRPS: ensemble forecasting using a model trained with a loss function based on the continuous ranked probability score},
  author={Lang, Simon and Alexe, Mihai and Clare, Mariana CA and Roberts, Christopher and Adewoyin, Rilwan and Ben Bouall{\`e}gue, Zied and Chantry, Matthew and Dramsch, Jesper and Dueben, Peter D and Hahner, Sara and others},
  journal={npj Artificial Intelligence},
  volume={2},
  number={1},
  pages={18},
  year={2026},
  publisher={Nature Publishing Group UK London}
}

@article{pinnington2026aifs,
  title={AIFS-DOP: End-to-End Medium-Range Weather Prediction from Observations Alone with Machine Learning},
  author={Pinnington, Ewan and Lean, Peter and Alexe, Mihai and Boucher, Eulalie and Lang, Simon and Laloyaux, Patrick and Mertes, Gert and Kral, Tomas and de Rosnay, Patricia and Chantry, Matthew and others},
  journal={arXiv preprint arXiv:2606.19093},
  year={2026}
}

@article{alet2026operational,
  title={Operational Tropical Cyclone Forecasting with AI},
  author={Alet, Ferran and Andersson, Tom R and Price, Ilan and Markou, Stratis and El-Kadi, Andrew and Masters, Dominic and Li, Amy and Merchant, Samier and Williams, Natalie and Thornton, Gregory and others},
  journal={Nature},
  pages={1--3},
  year={2026},
  publisher={Nature Publishing Group UK London}
}

@article{sonderby2020metnet,
  title={Metnet: A neural weather model for precipitation forecasting},
  author={S{\o}nderby, Casper Kaae and Espeholt, Lasse and Heek, Jonathan and Dehghani, Mostafa and Oliver, Avital and Salimans, Tim and Agrawal, Shreya and Hickey, Jason and Kalchbrenner, Nal},
  journal={arXiv preprint arXiv:2003.12140},
  year={2020}
}

@article{andrychowicz2023deep,
  title={Deep learning for day forecasts from sparse observations},
  author={Andrychowicz, Marcin and Espeholt, Lasse and Li, Di and Merchant, Samier and Merose, Alexander and Zyda, Fred and Agrawal, Shreya and Kalchbrenner, Nal},
  journal={arXiv preprint arXiv:2306.06079},
  year={2023}
}

@article{ni2025huracan,
  title={Huracan: A skillful end-to-end data-driven system for ensemble data assimilation and weather prediction},
  author={Ni, Zekun and Weyn, Jonathan and Zhang, Hang and Xiang, Yanfei and Bian, Jiang and Jin, Weixin and Thambiratnam, Kit and Zhang, Qi and Dong, Haiyu and Sun, Hongyu},
  journal={arXiv preprint arXiv:2508.18486},
  year={2025}
}

@article{gupta2026healda,
  title={HealDA: Highlighting the importance of initial errors in end-to-end AI weather forecasts},
  author={Gupta, Aayush and Subramaniam, Akshay and Pritchard, Michael S and Kashinath, Karthik and Frolov, Sergey and Lieberman, Kelsey and Miller, Christopher and Silverman, Nicholas and Brenowitz, Noah D},
  journal={arXiv preprint arXiv:2601.17636},
  year={2026}
}

@article{zhao2026skillful,
  title={Skillful high-resolution weather forecasting independent of physical models},
  author={Zhao, Pengcheng and Xiang, Siqi and Jin, Weixin and Ni, Zekun and Bian, Jiang and Fang, Zuliang and Sun, Hongyu and Zhang, Bin and Turner, Richard E and Weyn, Jonathan and others},
  journal={arXiv preprint arXiv:2605.28153},
  year={2026}
}

@article{zhao2024omg,
  title={OMG-HD: A high-resolution AI weather model for end-to-end forecasts from observations},
  author={Zhao, Pengcheng and Bian, Jiang and Ni, Zekun and Jin, Weixin and Weyn, Jonathan and Fang, Zuliang and Xiang, Siqi and Dong, Haiyu and Zhang, Bin and Sun, Hongyu and others},
  journal={arXiv preprint arXiv:2412.18239},
  year={2024}
}

@article{lam2023learning,
  title={Learning skillful medium-range global weather forecasting},
  author={Lam, Remi and Sanchez-Gonzalez, Alvaro and Willson, Matthew and Wirnsberger, Peter and Fortunato, Meire and Alet, Ferran and Ravuri, Suman and Ewalds, Timo and Eaton-Rosen, Zach and Hu, Weihua and others},
  journal={Science},
  volume={382},
  number={6677},
  pages={1416--1421},
  year={2023},
  publisher={American Association for the Advancement of Science}
}

@article{lean2023using,
  title={Using model cloud information to reassign low-level atmospheric motion vectors in the ECMWF assimilation system},
  author={Lean, Katie and Bormann, Niels},
  journal={Journal of Applied Meteorology and Climatology},
  volume={62},
  number={3},
  pages={361--376},
  year={2023}
}

@article{agrawal2025operational,
  title={An operational deep learning system for satellite-based high-resolution global nowcasting},
  author={Agrawal, Shreya and Hassen, Mohammed Alewi and Brempong, Emmanuel Asiedu and Babenko, Boris and Zyda, Fred and Graham, Olivia and Li, Di and Merchant, Samier and Potes, Santiago Hincapie and Russell, Tyler and others},
  journal={arXiv preprint arXiv:2510.13050},
  year={2025}
}

@inproceedings{gal2016dropout,
  title={Dropout as a bayesian approximation: Representing model uncertainty in deep learning},
  author={Gal, Yarin and Ghahramani, Zoubin},
  booktitle={international conference on machine learning},
  pages={1050--1059},
  year={2016},
  organization={PMLR}
}

@article{lavers2022evaluation,
  title={An evaluation of ERA5 precipitation for climate monitoring},
  author={Lavers, David A and Simmons, Adrian and Vamborg, Freja and Rodwell, Mark J},
  journal={Quarterly Journal of the Royal Meteorological Society},
  volume={148},
  number={748},
  pages={3152--3165},
  year={2022},
  publisher={Wiley Online Library}
}

@article{ferro2008effect,
  title={On the effect of ensemble size on the discrete and continuous ranked probability scores},
  author={Ferro, Christopher AT and Richardson, David S and Weigel, Andreas P},
  journal={Meteorological Applications: A journal of forecasting, practical applications, training techniques and modelling},
  volume={15},
  number={1},
  pages={19--24},
  year={2008},
  publisher={Wiley Online Library}
}

@article{liu2024global,
  title={Global-scale ERA5 product precipitation and temperature evaluation},
  author={Liu, Ronghua and Zhang, Xiaolei and Wang, Wei and Wang, Yun and Liu, Huageng and Ma, Meihong and Tang, Guoqiang},
  journal={Ecological Indicators},
  volume={166},
  pages={112481},
  year={2024},
  publisher={Elsevier}
}

@article{ingleby2015global,
  title={Global assimilation of air temperature, humidity, wind and pressure from surface stations},
  author={Ingleby, Bruce},
  journal={Quarterly Journal of the Royal Meteorological Society},
  volume={141},
  number={687},
  pages={504--517},
  year={2015},
  publisher={Wiley Online Library}
}

@article{smith2016multi,
  title={Multi-Radar Multi-Sensor (MRMS) severe weather and aviation products: Initial operating capabilities},
  author={Smith, Travis M and Lakshmanan, Valliappa and Stumpf, Gregory J and Ortega, Kiel L and Hondl, Kurt and Cooper, Karen and Calhoun, Kristin M and Kingfield, Darrel M and Manross, Kevin L and Toomey, Robert and others},
  journal={Bulletin of the American Meteorological Society},
  volume={97},
  number={9},
  pages={1617--1630},
  year={2016}
}

@article{moazami2021comprehensive,
  title={A comprehensive evaluation of GPM-IMERG V06 and MRMS with hourly ground-based precipitation observations across Canada},
  author={Moazami, SNMR and Najafi, MR},
  journal={Journal of Hydrology},
  volume={594},
  pages={125929},
  year={2021},
  publisher={Elsevier}
}

@article{glenn1950verification,
  title={Verification of forecasts expressed in terms of probability},
  author={Glenn, W Brier and others},
  journal={Monthly weather review},
  volume={78},
  number={1},
  pages={1--3},
  year={1950}
}

@article{murphy1977reliability,
  title={Reliability of subjective probability forecasts of precipitation and temperature},
  author={Murphy, Allan H and Winkler, Robert L},
  journal={Journal of the Royal Statistical Society Series C: Applied Statistics},
  volume={26},
  number={1},
  pages={41--47},
  year={1977},
  publisher={Oxford University Press}
}

@article{bauer2015quiet,
  title={The quiet revolution of numerical weather prediction},
  author={Bauer, Peter and Thorpe, Alan and Brunet, Gilbert},
  journal={Nature},
  volume={525},
  number={7567},
  pages={47--55},
  year={2015},
  publisher={Nature Publishing Group UK London}
}

@article{bougeault2010thorpex,
  title={The THORPEX interactive grand global ensemble},
  author={Bougeault, Philippe and Toth, Zoltan and Bishop, Craig and Brown, Barbara and Burridge, David and Chen, De Hui and Ebert, Beth and Fuentes, Manuel and Hamill, Thomas M and Mylne, Ken and others},
  journal={Bulletin of the American Meteorological Society},
  volume={91},
  number={8},
  pages={1059--1072},
  year={2010},
  publisher={American Meteorological Society}
}

@article{glahn1972use,
  title={The use of model output statistics (MOS) in objective weather forecasting},
  author={Glahn, Harry R and Lowry, Dale A},
  journal={Journal of Applied Meteorology and Climatology},
  volume={11},
  number={8},
  pages={1203--1211},
  year={1972},
  publisher={American Meteorological Society}
}

@article{gneiting2005calibrated,
  title={Calibrated probabilistic forecasting using ensemble model output statistics and minimum CRPS estimation},
  author={Gneiting, Tilmann and Raftery, Adrian E and Westveld III, Anton H and Goldman, Tom},
  journal={Monthly weather review},
  volume={133},
  number={5},
  pages={1098--1118},
  year={2005}
}

@article{hewson2021low,
  title={A low-cost post-processing technique improves weather forecasts around the world},
  author={Hewson, Timothy David and Pillosu, Fatima Maria},
  journal={Communications Earth \& Environment},
  volume={2},
  number={1},
  pages={132},
  year={2021},
  publisher={Nature Publishing Group UK London}
}

@article{vannitsem2021statistical,
  title={Statistical postprocessing for weather forecasts: Review, challenges, and avenues in a big data world},
  author={Vannitsem, St{\'e}phane and Bremnes, John Bj{\o}rnar and Demaeyer, Jonathan and Evans, Gavin R and Flowerdew, Jonathan and Hemri, Stephan and Lerch, Sebastian and Roberts, Nigel and Theis, Susanne and Atencia, Aitor and others},
  journal={Bulletin of the American Meteorological Society},
  volume={102},
  number={3},
  pages={E681--E699},
  year={2021}
}

@article{lang2025multi,
  title={A multi-scale loss formulation for learning a probabilistic model with proper score optimisation},
  author={Lang, Simon and Leutbecher, Martin and Maciel, Pedro},
  journal={arXiv preprint arXiv:2506.10868},
  year={2025}
}

@article{bi2023accurate,
  title={Accurate medium-range global weather forecasting with 3D neural networks},
  author={Bi, Kaifeng and Xie, Lingxi and Zhang, Hengheng and Chen, Xin and Gu, Xiaotao and Tian, Qi},
  journal={Nature},
  volume={619},
  number={7970},
  pages={533--538},
  year={2023},
  publisher={Nature Publishing Group UK London}
}

@article{kossaifi2026demystifying,
  title={Demystifying data-driven probabilistic medium-range weather forecasting},
  author={Kossaifi, Jean and Kovachki, Nikola and Mardani, Morteza and Leibovici, Daniel and Ravuri, Suman and Shokar, Ira and Calvello, Edoardo and Abbas, Mohammad Shoaib and Harrington, Peter and Subramaniam, Ashay and others},
  journal={arXiv preprint arXiv:2601.18111},
  year={2026}
}

@article{ravuri2021skilful,
  title={Skilful precipitation nowcasting using deep generative models of radar},
  author={Ravuri, Suman and Lenc, Karel and Willson, Matthew and Kangin, Dmitry and Lam, Remi and Mirowski, Piotr and Fitzsimons, Megan and Athanassiadou, Maria and Kashem, Sheleem and Madge, Sam and others},
  journal={Nature},
  volume={597},
  number={7878},
  pages={672--677},
  year={2021},
  publisher={Nature Publishing Group UK London}
}

@article {grecu2016gpm,
      author = "Mircea  Grecu and William S.  Olson and Stephen Joseph  Munchak and Sarah  Ringerud and Liang  Liao and Ziad  Haddad and Bartie L.  Kelley and Steven F.  McLaughlin",
      title = "The GPM Combined Algorithm",
      journal = "Journal of Atmospheric and Oceanic Technology",
      year = "2016",
      publisher = "American Meteorological Society",
      address = "Boston MA, USA",
      volume = "33",
      number = "10",
      doi = "10.1175/JTECH-D-16-0019.1",
      pages=      "2225 - 2245",
}

@inproceedings{alet2019graph,
  title={Graph element networks: adaptive, structured computation and memory},
  author={Alet, Ferran and Jeewajee, Adarsh Keshav and Villalonga, Maria Bauza and Rodriguez, Alberto and Lozano-Perez, Tomas and Kaelbling, Leslie},
  booktitle={international conference on machine learning},
  pages={212--222},
  year={2019},
  organization={PMLR}
}

\newpage
\appendix

\setcounter{section}{0}
\setcounter{figure}{0} 
\setcounter{table}{0}
\setcounter{equation}{0}
\renewcommand{\thesection}{A.\arabic{section}}
\renewcommand{\thefigure}{A.\arabic{figure}}
\renewcommand{\thetable}{A.\arabic{table}}
\renewcommand{\theequation}{A.\arabic{equation}}


{\noindent\LARGE\textbf{Appendix}\par}
\vspace{1em}

\section{Model details}
\label{sec:app_model}

\subsection{Model formulation}
\label{sec:app_model_formulation}
To restate our notation from \cref{sec:wn3}, the atmosphere at timestep $t$ and spatial location $s$ is represented by a set of $N$ spatio-temporal fields (indexed by $i$),
\begin{equation}
\label{eq:state}
\mathbf{W}_t = \bigl\{ f_i(s, t) : \mathcal{S}_i \times \mathcal{T}_i \to \mathcal{V}_i \bigr\}_{i=1}^{N},
\end{equation}
where each field $f_i$ denotes a physical variable over spatial domain $\mathcal{S}_i \subseteq \mathbb{S}^2$, continuous or discrete time subset $\mathcal{T}_i \subseteq \mathbb{R}$, and value space $\mathcal{V}_i$.

We parameterise the single-step conditional with a learnable operator~$\Phi_\theta$:
\begin{equation}
\label{eq:operator}
\mathbf{W}_{((t)\Delta,\;(t+1)\Delta]} = \Phi_\theta\!\left(\mathbf{W}_{((t{-}2)\Delta,\;t\Delta]},\; \boldsymbol{\xi}_t\right),
\end{equation}
where $\boldsymbol{\xi}_t$ is a single noise vector that parameterises aleatoric forecast uncertainty via conditional normalisation layers, and $(t)\Delta$ indexes the $t$-th $\Delta$-wide time interval. 

WN3 builds on FGN~\citep{alet2025skillful}, which underpins WeatherNext 2.

\paragraph{Latent grids and processor mesh.}
The operator~$\Phi_\theta$ follows an encode--process--decode structure. There are two input grids, one input at each resolution, indexed by $j \in \{0, 1\}$. Denote each input as $\mathcal{E}_{j,t} \in \mathbb{R}^{a_j \times b_j \times c_j}$ with spatial resolution $(a_j \times b_j )$ and  $c_j$ channels. $\mathcal{E}_{j,t}$ is first encoded point-wise to a latent grid $X_{j,t} \in \mathbb{R}^{a_j \times b_j  \times 1024}$, and then encoded by a Graph Neural Network (GNN) into a 6-times-refined icosahedral mesh latent $Z_{j,t}  \in \mathbb{R}^{40962 \times 1024}$, an operation which also produces an updated grid latent $X^{\text{skip}}_{j,t}  \in \mathbb{R}^{a_j \times b_j  \times 1024}$. A unified latent mesh $\mathbf{Z}_t = \sum_j Z_{j,t}$ is then processed by an $L$-layer graph transformer, before being decoded back to the latent grid and added to $X^{\text{skip}}_{j,t}$ yielding $\bar{X}_{j,t}$.

\paragraph{Gridded and continuous output heads.}
Gridded fields are decoded onto fixed latitude-longitude grids at their native resolutions $(a_j \times b_j)$, pointwise, from $\bar{X}_{j,t}$ via a Multilayer Perceptron (MLP). Surface station predictions are decoded as a continuous query from a \tenthdegree latent grid (i.e. $a_j = 1801,  b_j=3600$): for any point $(s, \delta t) \in \mathcal{S}^2 \times [0, \Delta]$, the prediction is
\begin{equation}
\label{eq:station}
\mathbf{v}(s, \delta t) = \text{MLP}_{\text{stn}}\!\left(\,\text{Interp}\!\left(\psi([X_{j,t}^{\text{skip}} \,\|\, \bar{X}_{j,t}]),\, s\right) \;\|\; \mathbf{m}(s, \delta t)\,\right),
\end{equation}
where $\text{Interp}(\cdot, s)$ bilinearly interpolates a grid latent to $s$ after the latent is processed by a 4-layer, width-768 convolutional neural network (CNN)~$\psi$. The input to this CNN is the concatenation of the input and decoded latents $[X_{j,t}^{\text{skip}} \,\|\, \bar{X}_{j,t}]$, $\mathbf{m}(s, \delta t) = \text{MLP}_{\text{mdata}}([\text{elevation}(s), \text{is\_land}(s), \delta t])  \in \mathbb{R}^{768}$ encodes local surface features alongside the normalized relative observation query time $\delta t \in [-1, 0]$ with a 1-hidden layer width-768 MLP, $\text{MLP}_{\text{stn}}$ is a 2-hidden-layer width-768 MLP, and $\|$ indicates concatenation along the feature axis.  This follows the continuous querying approach of \citet{alet2019graph} from a common processor mesh, enabling training on sparse, irregularly located observations and inference at arbitrary spatial and temporal query points.

\paragraph{Training objective.}
The model is trained end-to-end on a composite loss accumulated over $R$~autoregressive rollout steps. We use the fair Continuous Ranked Probability Score (CRPS) with two sampled trajectories ~$f(\cdot,\xi^1_{1:t}),f(\cdot,\xi^2_{1:t})=:f^{1,2}$ and corresponding target $\hat{f}$. For field $i$, the loss $\ell_i(\cdot, \cdot)$~ averages over that field's own set of spatial positions~$\mathcal{S}_i \subset \mathcal{S}^2$ and temporal samples~$\mathcal{T}_i$ , 
so that the per-modality weights~$\lambda_i$ (see Table \ref{tab:wn3_loss_weights}) control relative importance independently of grid size or sampling density:
\begin{equation}
\label{eq:loss} 
\mathcal{L} = \sum_{i=1}^{N}\, \lambda_i \;\underset{(s,\, q) \,\in\, \mathcal{S}_i \times \mathcal{T}_i}{\mathrm{avg}}\; \ell_i\!\left(\hat{f}_i(s, q),\; f_i^{1,2}(s, q)\right).
\end{equation}
For stations, the average runs over the variable-sized set of available reports within each $\Delta$-wide window. For gridded cyclone output fields, the average runs over all grid points for which the gridded target is not NaN (see \cite{alet2026operational}). This design ensures that sparse modalities (stations, cyclones) are not overwhelmed by the much larger gridded fields.

Within the atmospheric analysis fields, each variable--pressure-level pair~$(v, p)$ carries its own weight~$\lambda_{v,p}$:
\begin{equation}
\label{eq:loss_atm}
\mathcal{L}^{\text{atm}}_t = \sum_{v=1}^{C} \sum_{p=1}^{P} \lambda_{v,p} \; \underset{(s, h) \in \mathcal{S}_v \times \mathcal{T}_v}{\operatorname{avg}} \; \text{CRPS}\!\left(\hat{f}_{v,p}(s, h), \, f_{v,p}^{1,2}(s, h)\right).
\end{equation}

where pressure level weighting is as in \citet{alet2026operational} and relative variable weightings are specified in Table \ref{tab:wn3_loss_weights}. For all variables specified as input and output in Table \ref{tab:data}, with the exception of geostationary satellite fields, the target above is the residual with respect to the last input frame. Similarly, analysis variables defined on subtime channels target the residual with respect to the corresponding non-subtime variable on the outer frames. This includes the pressure-level slices, for which the corresponding level from the outer frame is used.

\subsection{Additional model modifications}
Apart from the key features mentioned in the main text, we add the following model changes:
\paragraph{Global mean loss.}

For precipitation and cloud cover variables we observed that models trained solely on local (point-wise) marginal CRPS exhibited substantial per-sample bias across individual ensemble members. To mitigate this, we incorporated a globally pooled CRPS component into the training loss, weighted at a factor of $0.3$ to the corresponding base variable's loss weight. We found this to be a simple-but-effective fix, related to spherical harmonic losses~\citep{kochkov2024neural} or Gaussian smoothing~\citep{lang2025multi}. While this pooled objective effectively eliminates per-sample bias for some variables, we found that applying it to sparse station variables exacerbated mesh-scale artifacts; we therefore omitted the pooled loss for station targets.

\paragraph{Cumulative variables.}
WN3 uses \hresfc as both an input and a training target. We model cumulative variables (\textit{cdir}, \textit{fdir}, \textit{ssrd} and \textit{tp}) as the accumulation within a given time period (e.g. for an initialization time of 03 UTC and a lead time of 1 hour it would be the accumulation from 03 UTC to 04 UTC). In HRES these variables are defined as accumulations from time 0, so we "de-accumulate" them. However, for \hresfczero this leaves us with a constant value of 0. In this case, during training, we populate these variables using the HRES-fc6 values from the prior initialization time as necessary. For these cumulative variables, we predict both 1-hour and 6-hour accumulations.

\paragraph{Input dropout for operational robustness.}
Several autoregressive inputs are unavailable on the first forecast step: as mentioned previously, cumulative analysis variables have a value of 0 in \hresfczero, while both IMERG and PARDIG products are currently not available operationally with an acceptable latency. Similarly, for older training years, some inputs (geostationary satellite imagery, IMERG and PARDIG) do not exist; in these cases, the corresponding input channels are filled with their training-set mean, to mask them out. To bridge the train--inference gap for autoregressive inputs, we apply stochastic input dropout during training: with probability~$0.9$ an input modality is masked as if it were the first rollout step (unavailable); with probability~$0.02$, as the second step (partially available from the previous output); and with the remaining $0.08$~probability, the full input is provided (falling back to prior HRES-fc6 values for \hresfczero cumulative variables, as described above). This teaches the model to produce skilful forecasts regardless of which autoregressive inputs are available, removing the need for separate handling at initialization time.

\paragraph{Sea surface temperature (\textit{sst}).} As in \cite{alet2026operational}, missing values over land in ERA5 are replaced with the minimum value of the variable seen in a subset of the dataset. For \hresfc, we imputed this same minimum value over land for \textit{sst}, using ERA5's SST validity mask at $0.25^\circ$, and HRES's land sea mask at $0.1^\circ$, to define the land mask.

\paragraph{Spatial sharding.}
Given the scale of WN3, we introduce spatial sharding of the processor mesh: the icosahedral mesh nodes are partitioned along a spatial axis (longitude by default), and gather/scatter operations across grid-to-mesh and mesh-to-grid edges are executed as local operations within each partition, with only the mesh-to-grid scatter requiring global communication. Sorting the grid-to-mesh edges by senders makes the most expensive sharding operation local, substantially reducing cross-device communication.

\subsection{Training}
\label{sec:app_training}

WN3 is trained via a multi-stage curriculum that progressively increases spatial resolution and introduces additional observation modalities. \cref{tab:training} summarises the full stage sequence and \cref{tab:wn3_hyperparams} lists hyperparameters used for training.

\begin{table}[th]
\centering
\scriptsize
\setlength{\tabcolsep}{3pt} 
\begin{tabular}{c | l | c | >{\raggedright\arraybackslash}p{3.05cm} | >{\raggedright\arraybackslash}p{3.05cm} | c | c}
\toprule
\# & Stage & $\Delta$ & "Coarse" Grid & "Fine" Grid & Stations & Backbone \\
\midrule
1 & $1^\circ$ pretraining (12h)    & 12h & \multicolumn{2}{p{6.3cm}|}{$1^\circ$. $<2016$: ERA5; $<2023$: \hresfc, Geosats, IMERG, PARDIG} & - & Train \\
2 & $1^\circ$ pretraining (6h)     & 6h  & \multicolumn{2}{c|}{Same as Stage 1} & - & Train \\
3 & $0.25^\circ$ pretraining       & 6h  & \multicolumn{2}{c|}{$0.25^\circ$. Same as Stage 1} & - & Train \\
4 & Cyclone warmup                 & 6h  & \multicolumn{2}{c|}{Same as Stage 3 $+$ Cyclones $<2023$} & - & \textbf{Frozen} \\
5 & Cyclone e2e                    & 6h  & \multicolumn{2}{c|}{Same as Stage 4} & - & Train \\
6 & $0.1^\circ$ pretraining        & 6h  & $0.25^\circ$. $<2023$: \hresfc, Cyclones, Geosats, IMERG, PARDIG & $0.1^\circ$. $<2023$: \hresfc, Geosats, IMERG, PARDIG & - & Train \\
7 & AR finetuning ($2{\to}7$)      & 6h  & Same as Stage 6 & Same as Stage 6 & - & Train \\
8 & Frozen finetuning (stations) 1 & 6h  & Same as Stage 6 & Same as Stage 6 & $<2023$ METAR & \textbf{Frozen} \\
9 & AR finetuning (8)              & 6h  & Same as Stage 6, extended to $<2024/26$ & Same as Stage 6, extended to $<2024/26$ & - & Train \\
10 & Frozen finetuning (stations) 2 & 6h & Same as Stage 9 & Same as Stage 9 & $<2024/26$ METAR & \textbf{Frozen} \\
\bottomrule
\end{tabular}
\caption{\textbf{Training stages for WeatherNext~3} (for data start dates, see \cref{tab:data}).}
\label{tab:training}
\end{table}

\begin{table}[htp]
\centering
\small
\begin{tabular}{rlccccccc}
\toprule
\# & Stage & Peak LR & Total Steps & Gradient Norm Clipping \\
\midrule
1  & 1$^{\circ}$ pretraining (12h)     & 4.95e-4                      & 400 000 & 3.5  \\
2  & 1$^{\circ}$ pretraining (6h)      & 7.07e-5                      & 100 000 & 5.5  \\
3  & 0.25$^{\circ}$ pretraining        & 7.07e-5                      &  32 000 & 6.5  \\
4  & Cyclone warmup                    & 7.07e-5                      &   5 000 & 6.5  \\
5  & Cyclone e2e                       & 7.07e-5                      &  20 000 & 6.5  \\
6  & 0.1$^{\circ}$ pretraining         & 7.07e-5                      &   8 000 & 5.5  \\
7  & AR finetuning (2$\to$7)           & 7.07e-6, [7.07e-7]$\times$5 &   9 000 & 11.0 \\
8  & Frozen finetuning (stations) 1    & 7.07e-5                      &  12 000 & 5.5  \\
9  & AR finetuning (8)                 & 7.07e-7                      &   1 000 & 11.0 \\
10 & Frozen finetuning (stations) 2    & 1.17e-5                      &   4 000 & 5.5  \\
\bottomrule
\end{tabular}
\caption{\textbf{Training hyperparameters for WeatherNext~3 stages}. Across all stages: the optimizer is AdamW with weight decay 0.1 and $\beta_2 = 0.95$; a cosine decay is used for the learning rate schedule, with warmup steps $\min(1000, 0.1 * \text{total steps})$; the batch size is 64 and the number of training samples in the FGN setup is 2.}
\label{tab:wn3_hyperparams}
\end{table}

\begin{table}[t]
\centering
\small
\begin{tabular}{lcccc}
\toprule
& \multicolumn{2}{c}{\textbf{"Coarse" grid}} & \multicolumn{2}{c}{\textbf{"Fine" grid}} \\
\cmidrule(lr){2-3} \cmidrule(lr){4-5}
\textbf{Target Variables} & \textbf{Main (6h)} & \textbf{Subtime (1h)} & \textbf{Main (6h)} & \textbf{Subtime (1h)} \\
\midrule
\textit{t}, \textit{u}, \textit{v}, \textit{q}, \textit{w} & 1.0 & - & - & - \\
\textit{z} & 0.2 & - & - & - \\
300/500 hPa \textit{t}/\textit{z}; 1000 hPa \textit{u}/\textit{v} & - & 0.1 & - & - \\
\textit{2t} & 0.5 & 0.5 & 0.5 & 0.5 \\
\textit{10u}, \textit{10v}, \textit{100u}, \textit{100v} & 0.025 & 0.025 & 0.025 & 0.025 \\
\textit{2d}, \textit{msl}, \textit{sst}, \textit{tcc}, \textit{hcc}, \textit{mcc}, \textit{lcc}, \textit{tp\_1hr} & 0.05 & 0.05 & 0.05 & 0.05 \\
\textit{tp\_6hr} & 0.05 & - & 0.05 & - \\
\textit{ssrd\_1hr}, \textit{cdir\_1hr}, \textit{fdir\_1hr} & $1/60$ & $1/60$ & $1/60$ & $1/60$ \\
\textit{ssrd\_6hr}, \textit{cdir\_6hr}, \textit{fdir\_6hr} & $1/60$ & - & $1/60$ & - \\
\midrule
PARDIG 1h rain rate & 0.75 & - & 0.75 & - \\
IMERG 1h rain rate & 0.543 & - & 0.543 & - \\
Geosats mosaic (each 11 spectral channels) & 0.05 & - & 0.05 & - \\
Cyclone existence, $V_{\max}$, $P_{\min}$ & 20.0 & - & - & - \\
\parbox[t]{5.5cm}{Cyclone structure: $R_{\max}$, \newline 34/50/64-kt quadrant radii} & 2.0 & - & - & - \\
\bottomrule
\end{tabular}
\caption{\textbf{Per-variable loss weights across spatial and temporal representations for WeatherNext~3.} Hyphens (-) indicate that the variable is not evaluated on that representation. Sparse METAR variables \textit{2t}, \textit{2d} are weighted at $1.0$.}
\label{tab:wn3_loss_weights}
\end{table}

\paragraph{Progressive increase of resolution.}
The model begins training on ERA5 ($<$2016)/\hresfc ($\geq 2016$) with geostationary satellite imagery, IMERG and PARDIG precipitation, and hourly sub-step analysis targets. This is initially at $1^\circ$ resolution with $12\,\text{h}$ outer steps. This then decreases to $6\,\text{h}$ steps, after which resolution is progressively increased to $0.25^\circ$ and finally $0.1^\circ$. The model is trained with two grids: "coarse" and "fine", both of which are encoded and decoded (supporting variables as input and output) at each step. All gridded variables are modelled by the coarse grid, while the fine grid separately only models the subset of variables ultimately available at $0.1^\circ$ resolution. In $1^\circ$ and $0.25^\circ$ stages, both grids operate at the respective stage's resolution. In the final $0.1^\circ$ stages, the "coarse" grid remains at $0.25^\circ$ while the "fine" grid is increased to $0.1^\circ$. Each grid resolution increase is accompanied by a scaling of the sum of message vectors in the encoder GNN to preserve the scale of the incoming signal. From $1^\circ \rightarrow 0.25^\circ$ this is $\frac{1}{4}$ and from $0.25^\circ \rightarrow 0.1^\circ$ this is $\frac{1}{25}$.

\paragraph{Normalization.}
All input variables are normalized to unit variance and zero mean. Targets (in their potentially residual form) are similarly normalized, with the exception of the cyclone existence field and subtime analysis variables. The former is normalized by definition, while the latter are approximately nomalized using the statistics of the outer-frame differences.

\paragraph{Cyclone finetuning.}
Cyclone attribute maps are introduced with a two-step process: first a \emph{warmup} stage trains only the newly added output heads with the backbone frozen, then an \emph{end-to-end} stage unfreezes the full model for joint optimisation.

\paragraph{Autoregressive finetuning.}
The model is finetuned with progressively longer rollouts, from 2 to 8 autoregressive steps. These phases use all previously introduced modalities as both input and target.

\paragraph{Frozen finetuning at high resolution.}
The station observation heads are trained while keeping the backbone and all other heads fixed at $0.1^\circ$ resolution. Frozen finetuning is interleaved with AR finetuning via a schedule: an initial frozen stage (75\% of total steps) after 7AR, then the final 8AR stage, followed by a second frozen stage (25\% of steps) that merges the final backbone weights from AR with the station weights from the initial frozen stage. 

\paragraph{Data recency.}
The interleaving process above is done to enable finetuning up to different years, in a compute efficient manner, while maintaining causality for the different evaluations. Up to the end of the first frozen finetuning stage, the model is trained with data spanning up to the end of 2022. At 8AR, more recent data is introduced (and so is also used in the final stage of frozen finetuning). We use this to generate "splits" of the model, each trained with data up to the start of its corresponding year: <2023 (used for validation in research process), <2024 (for non-realtime evaluations), <2025, <2026 and <2026-07-01 (for the production model and realtime evaluations). For the latter two splits, IMERG Final is only available up to September 2025 so we employ the mean-masking described above for the variable's entire frame.

\paragraph{Output clipping.}
We noticed that in inference, some stationary pixels have increasingly unrealistic values during rollout. We traced this back to pixels going out of the training distribution, specifically for bounded variables like cloud cover. As a simple fix we clip those bounded variables during inference ($[0, 1]$  for cloud cover, $\geq 0$ for specific humidity, solar radiation and precipitation) at each step, before feeding the output auto-regressively. This simple solution prevents the occurrence of those pixel artifacts.

\paragraph{Computational resources.}
Each model seed took roughly 6.7 days of wall clock time to train. The exact hardware configuration varied depending on training stage. In total, both models required approximately 4.8 chip-years on TPUv4 and 8.9 chip-years on TPU7x chips. Generating a single ensemble member forecast takes roughly 6.3 minutes of wall clock time using 4 TPUv5p chips.

\section{Additional information on evaluation}
\label{sec:app_eval}

\subsection{CRPS}\label{sec:app_eval:crps}
We use the biased version of the CRPS because WN3 as a multi-seed ensemble violates the assumption of i.i.d. inherent in the unbiased CRPS \citep{ferro2008effect}. Further, this more accurately reflects the skill of the forecasts users would actually receive operationally rather than that of a theoretical ensemble of infinite size. We make one exception for our ENS baselines, where due to various constraints we did not have the full 50+1 member ensemble available. We note, however, that the differences between biased and unbiased metrics for ensemble sizes of $O(50)$ are small.

\subsection{Pooled CRPS}\label{app:pooled_crps}
Pooled CRPS computes CRPS on fields which have undergone dimensionality-preserving spatial average- or max- pooling. Pooling regions are defined on the latitude-longitude grid to be approximately equi-area patches, defined by their size in degrees at the equator (and thus warping as latitude increases or decreases towards the poles). A pooled value is computed for a patch centred at every grid point, and so latitude-based weighting is applied when computing the overall spatial average of the pooled CRPS, to account for over-counting at higher/lower latitudes. 

\section{Additional information on datasets}
\label{sec:app_data}

\begin{table}[bh]
\centering
\small
\label{tab:var_abbrev_single}
\begin{tabular}{@{} l l @{}}
\toprule
\textbf{Short name} & \textbf{Long name} \\
\midrule
\textit{z} & Geopotential \\
\textit{t} & Temperature \\
\textit{q} & Specific humidity \\
\textit{u} & Zonal wind component \\
\textit{v} & Meridional wind component \\
\textit{w} & Vertical velocity \\
\textit{2t} & 2\,m temperature \\
\textit{2d} & 2\,m dewpoint temperature \\
\textit{10u} & 10\,m zonal wind component \\
\textit{10v} & 10\,m meridional wind component \\
\textit{100u} & 100\,m zonal wind component \\
\textit{100v} & 100\,m meridional wind component \\
\textit{msl} & Mean sea level pressure \\
\textit{tcc} & Total cloud cover \\
\textit{hcc} & High cloud cover \\
\textit{mcc} & Medium cloud cover \\
\textit{lcc} & Low cloud cover \\
\textit{cdir*} & Clear-sky direct solar radiation at surface \\
\textit{fdir*} & Total-sky direct solar radiation at surface \\
\textit{ssrd*} & Surface solar radiation downwards \\
\textit{tp*} & Total precipitation \\
\textit{tisr*} & Total incident solar radiation \\
\bottomrule
\end{tabular}
\caption{List of variable abbreviations and full descriptions. * variables are cumulative and may be denoted with an accumulation period suffix (\textit{\_1hr}, \textit{\_6hr}) where relevant.}
\end{table}

Here we provide additional details for some of the datasets already mentioned in the main text.

\subsection{ECMWF ENS and AIFS ENS baselines}
\label{sec:app_ens}
For our 2024 evaluations, we download upper-level ENS forecasts for 00/12 initializations from TIGGE \citep{bougeault2010thorpex} in the native grid and interpolate it to \quarterdegree. For single-level variables, we download native grid TIGGE forecasts for 2t, 2d and 10u/v for 6 hour lead time multiples. For the hourly lead times in-between we download native grid data from MARS. For all other single level variables, we download all lead times from MARS. We then interpolate those native grid data to \tenthdegree. However, we noticed that there seems to be a difference between the forecasts downloaded from TIGGE and MARS, stemming from the fact that TIGGE forecasts are stored at a reduced resolution\footnote{https://confluence.ecmwf.int/spaces/TIGGE/pages/40109876/Models}. This is evident in the hourly evaluations (\cref{fig:fig_appendix_0p1_analysis_scores_48h} and \cref{fig:fig_appendix_hourly_station_evals_48h}). Since our ground truth is entirely downloaded from MARS, these discrepancies cause small spikes in error for those variables at 6h lead time multiples. Without further analysis, we can only estimate the magnitude of those spikes from the shape of the curves. For 2m temperature, which is the most affected variable, the effect is roughly 0.02K in CRPS for \tenthdegree analysis evaluations (note that the 6-hourly step pattern in \cref{fig:fig_appendix_0p1_analysis_scores_48h} is predominantly caused by another mechanism, discussed in the figure caption). Since the gap to WN3 is generally substantially larger than 0.02K, these discrepancies should not qualitatively affect the results. In \cref{fig:figure_2_analysis}, the only impacted variable is 2m dewpoint temperature. For station evaluations, 10m wind speed seems to be most affected, with no big difference in 2m temperature and relative humidity. For wind speed the difference is on the order of 0.02 m/s CRPS, which is much smaller than the gap to WN3. 

Also note that, for download efficiency, we downloaded only 48 members for ENS (without the control forecast). We use the fair CRPS for ENS \tenthdegree in all evaluations to account for this difference. 

For the real-time 2026 evaluations we obtain all data from the ECMWF open-data program\footnote{https://www.ecmwf.int/en/forecasts/datasets/open-data}. The ENS data there is only provided at \quarterdegree. For AIFS ENS we use the AIFS ENS control member at $t=0$ as a ground truth. This data stems from the same raw fields as the \hresfczero we use but is regridded differently. While we directly convert from the native O1280 reduced Gaussian grid to \quarterdegree, for AIFS the fields are first regridded to the AIFS-native O320 grid, before being interpolated to \quarterdegree. This introduces differences that affect scores meaningfully.

\subsection{Geostationary satellite mosaic.}

We construct our own real-time mosaic of geostationary satellite imagery by combining data streams of several satellites, following the methodology in \cite{agrawal2025operational}. The composite spans 11 spectral channels across visible and infrared regimes (central wavelengths at $0.64$, $0.863$, $1.61$, $3.83$, $6.20$, $7.33$, $8.59$, $9.62$, $11.20$, $12.40$, and $13.30\mu\text{m}$). While the native spatial resolution varies across instruments and channels, we regrid them all to \tenthdegree for training WN3. The historical dataset extends back to 2016 and covers several generations of satellites and instruments; the real-time configuration currently uses the GOES-18/19, Meteosat-9/11, and Himawari-8/9 satellites.

\subsection{In-situ surface observations}

\begin{figure}[h]
    \centering\includegraphics[width=1.0\linewidth]{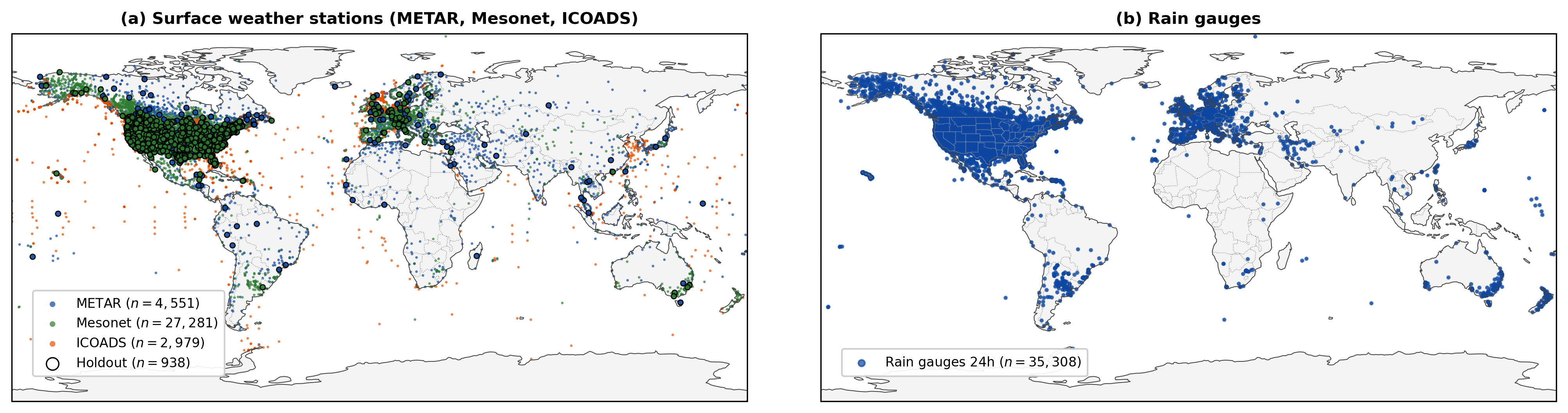}
    \caption{Distribution of in-situ observations for 1 January 2024 00 UTC. (a) METAR, Mesonet and ICOADS stations. Hold-out stations with outline. (b) Rain-gauge observations.}
    \label{fig:fig_a1_station_distributions}
\end{figure}

We build a combined hourly dataset of METAR, Mesonet (available from MADIS) and ICOADS \citep{icoads_release3_dataset} for training starting in June 2001. For Mesonet, which provides some sub-hourly observations, if more than one observation is available per hour for a given station, we pick the observation closest to the hour. For METAR and Mesonet we use the quality control flags provided with the data to filter out bad-quality observations. For ICOADS, no such flags are provided. However, we noticed clear outliers. To remove those, we first enforce rough physical bounds: -20 to 40$^\circ\text{C}$ for temperature and dewpoint, and 0 to 50 $m/s$ for wind speed. Second, we then apply a departure check from ERA5 for all three datasets, where data points that deviate more than 5$^\circ\text{C}$, or 5 $m/s$ are removed. This only applies to training. For evaluation, we only remove data based on the provided quality control flags. We hold out 5\% of METAR and Mesonet stations randomly from training, which we then use for independent evaluation.

\subsection{Elevation and land-sea mask}
To make predictions using the trained station head at any location in the globe, we need to provide the appropriate metadata: latitude, longitude, elevation and land/sea. For elevation, we create a global \twentiethdegree elevation map by reprojecting the Global Multi-resolution Terrain Elevation Data \citep{danielson2011gmted2010}. For the land-sea mask we use the ESA WorldCover 10m v200 dataset \citep{zanaga2022worldcover2021}. The reprojection is done by taking the mean of the high-resolution datasets, so that each \twentiethdegree grid point represents the grid box average.

\subsection{IMERG}
NASA’s Integrated Multi-satellitE Retrievals for GPM (IMERG) product \citep{huffman2023nasa} (IMERG) is a widely used precipitation estimate based on microwave and infrared observations from the GPM constellation. Specifically, we use IMERG Final, which has been additionally calibrated against gauge observations. IMERG is available at \tenthdegree spatial and 30\,minute temporal resolution, which we sum to hourly accumulations for model training. Note that the \tenthdegree grid used by IMERG and WN3 are offset by \twentiethdegree longitudinally, requiring us to linearly interpolate the dataset. This is only done for training. For evaluation we interpolate all forecasts to the native IMERG grid. We use IMERG starting in 2000. Due to IMERG's transition to v08, IMERG Final is currently only available until September 2025\footnote{https://gpm.nasa.gov/data/news/imerg-v08-transition-schedule}. 


\subsection{PARDIG}
We developed our own global precipitation estimate, called PARDIG (Precipitation AI Reanalysis - Densely Inferred GPM). This is based on a separately trained model with architecture similar to WN3, albeit with no noise injection. The input to this model is a 4 hour window of geostationary satellite data, ERA5/HRES pressure level and surface variables, and a mosaic of microwave sounder data; the target of the model is the GPM CORRA measurements \citep{grecu2016gpm} at the two hour window centred within the input window. The CORRA product uses a combination of microwave and radar measurements from the core observatory satellite in the GPM program. This satellite is in low Earth orbit and thereby provides sparse data points. The precipitation estimation model uses a similar sparse output head architecture that is used for WN3's station head (targeting a binned probability distribution). After training, this enables us to make predictions globally. We create \tenthdegree predictions of instantaneous rain rate every 15 minutes, and then sum those to produce an hourly dataset starting in 1999, which then is used as a target for WN3. In preliminary evaluations we find this precipitation estimate to be substantially more skillful when evaluated against ground-based radar or rain gauges compared to IMERG Final. Further details will be provided in an upcoming paper.

\subsection{MRMS}
For evaluation, we also use the Multi Radar Multi Sensor (MRMS) Quantitative Precipitation Estimate Pass-2 product for independent evaluation \citep{noaa_nssl_mrms}. This provides hourly precipitation estimates derived mainly from radar but calibrated against rain gauges. We exclude areas with a low radar quality index for evaluation to ensure we are using the highest quality data. We further restrict the evaluation to a spatial domain spanning latitudes $24^\circ\text{N}$ to $50^\circ\text{N}$ and longitudes $126^\circ\text{W}$ to $65^\circ\text{W}$.

\subsection{IBTrACS gridded fields}
\label{app:cyclones}
We follow the approach used in \cite{alet2026operational} to convert IBTrACS data \citep{knapp2010ibtracs,knapp2024ibtracs} into gridded targets for WN3, with two small modifications.
\cite{alet2026operational} used an imputation approach for the wind quadrant radii: for each wind threshold (34kt, 50kt and 64kt), if at least one quadrant radius is not NaN, the rest of the quadrant radii were imputed to the mean of the available quadrant radii.
If all of them were NaN, they were all set to zero.
We modified this approach, by removing the mean-imputation and only setting a NaN to a zero if the maximum sustained wind speed of the cyclone is below the corresponding wind threshold, i.e. if we can unambiguously determine that the NaN represents a zero quadrant radius.
In addition, we perform temporal interpolation on the IBTrACS dataset, to bring it to a 1-hourly resolution, before gridding it.
To interpolate the cyclone position, we perform the linear interpolation in 3D Euclidean coordinates and project back to longitude and latitude, while performing regular linear interpolation for the rest of the scalar variables.
Finally, the interpolation may introduce unphysical combinations of wind speeds and quadrant radii, e.g. if at an interpolated time the wind speed drops below 34kt, then all 34kt radii should be set to zero.
This happens relatively infrequently but we nonetheless account for it by enforcing physical consistency on the ground truth 1-hourly IBTrACS data, following the approach outlined in \cite{alet2026operational}.

\newpage

\begin{figure}[h]
    \centering\includegraphics[width=1.0\linewidth]{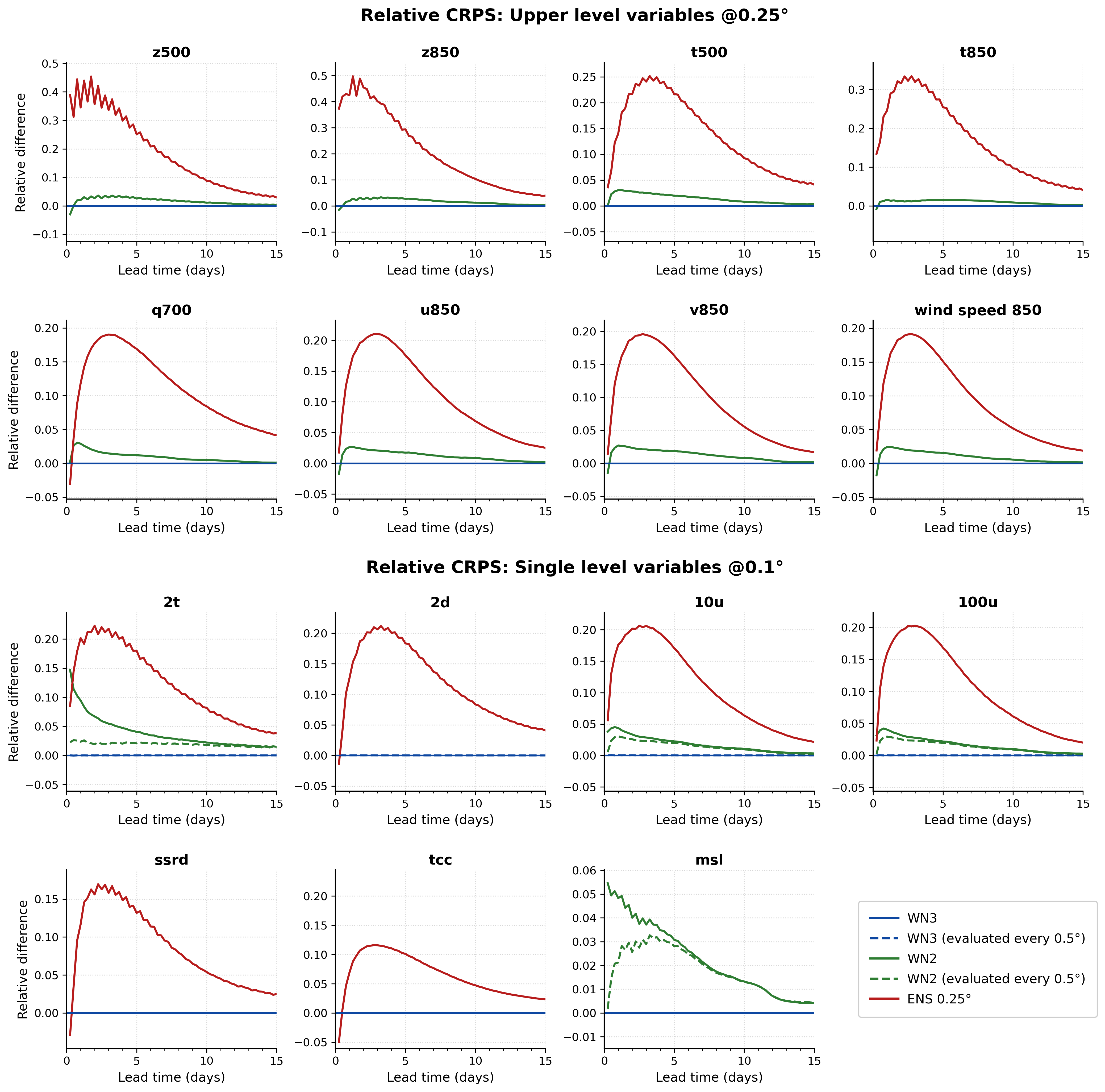}
    \caption{Relative CRPS difference with respect to WN3. (top) Variables evaluated at \quarterdegree. (bottom) Variables evaluated at \tenthdegree. Additionally, WN3 and WN2 evaluated at common grid points (\halfdegree; dashed lines) to separate the effect of regridding WN2 from \quarterdegree to \tenthdegree from pure model skill. Note that for WN3, the differences are indiscernible. For WN2 the largest differences are for 2t and msl at early lead hours, indicating that there is a substantial penalty from regridding. However, even when removing this penalty, WN3 still outperforms WN2 on those variables. Note that not all variables are available for WN2 and we did not download msl for ENS. 00/12 UTC initializations for 2024.}
    \label{fig:fig_appendix_analysis_relative_lines_2024}
\end{figure}

\begin{figure}[h]
    \centering\includegraphics[width=1.0\linewidth]{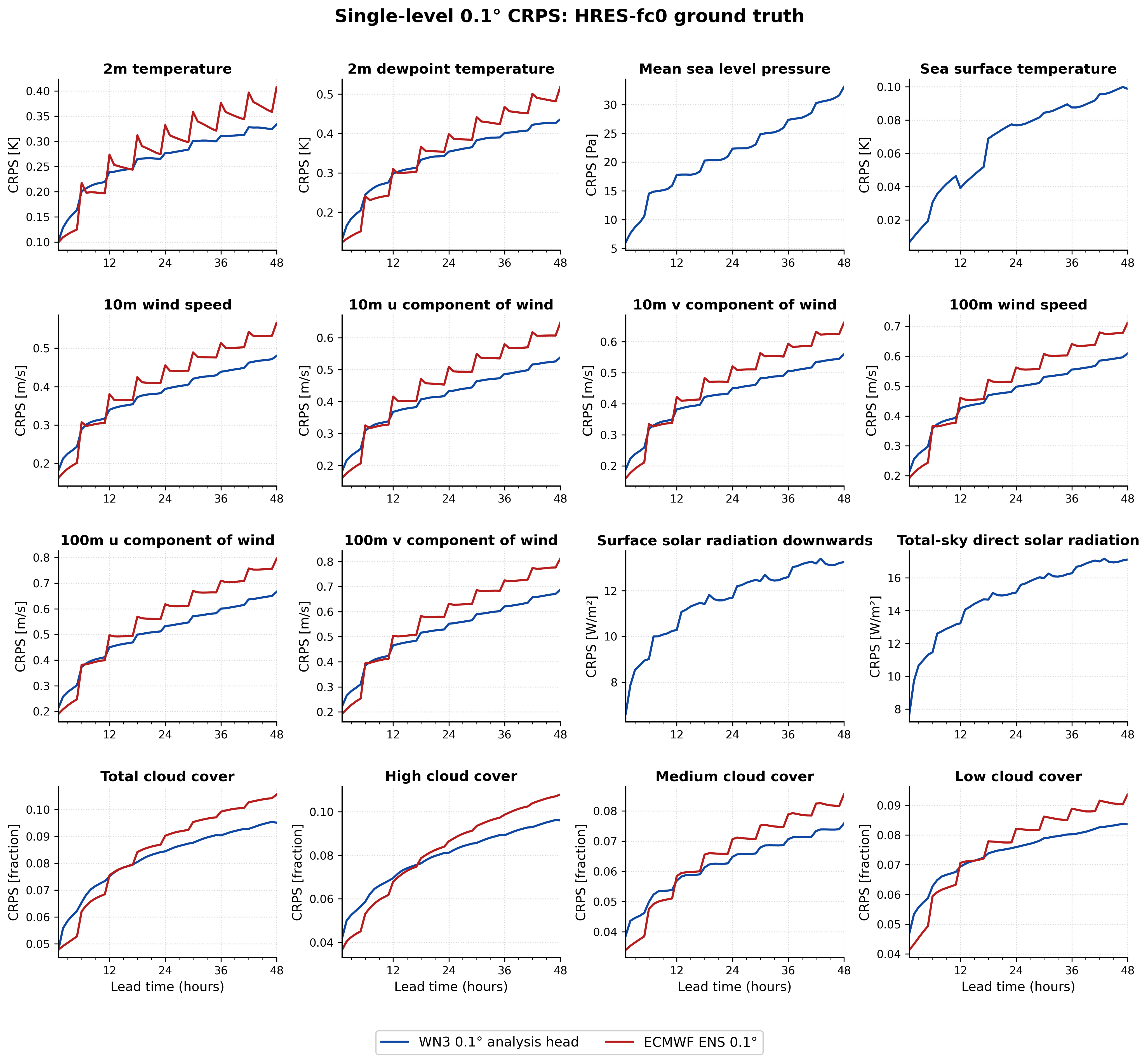}
    \caption{Hourly CRPS for single-level variables for the first 48h lead time with \hresfc ground truth. The general step shape of the plots, with a substantial step change every 6 hours, is explained by the fact that the ground truth goes from being a continuous HRES forecast trajectory for the first 5 hours to a new analysis state at the next 6h multiple. For 2m temperature, 2m dewpoint, and all wind variables, there is an additional contribution to the spike at 6h multiples which is explained in \cref{sec:app_ens}. We did not have ENS data available for some variables. 00/12 UTC initializations for 2024.}
    \label{fig:fig_appendix_0p1_analysis_scores_48h}
\end{figure}

\begin{figure}[h]
    \centering\includegraphics[width=1.0\linewidth]{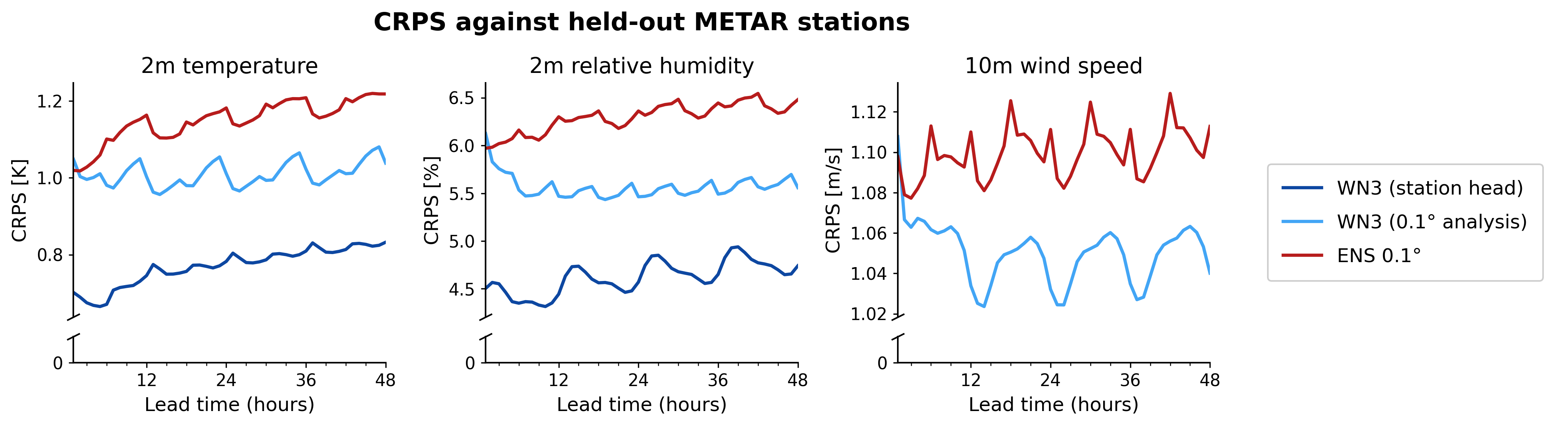}
    \caption{Hourly station evaluations against held-out METAR stations. ENS exhibits some spikes in CRPS at 6 hour intervals as discussed in \cref{sec:app_ens}. WN3 \tenthdegree analysis predictions show an initially large CRPS which is likely due to a mismatch between the initial conditions and the station observations. As ensemble spread is introduced, CRPS sharply improves. 00/12 UTC initializations for 2024.}
    \label{fig:fig_appendix_hourly_station_evals_48h}
\end{figure}

\begin{figure}[h]
    \centering\includegraphics[width=1.0\linewidth]{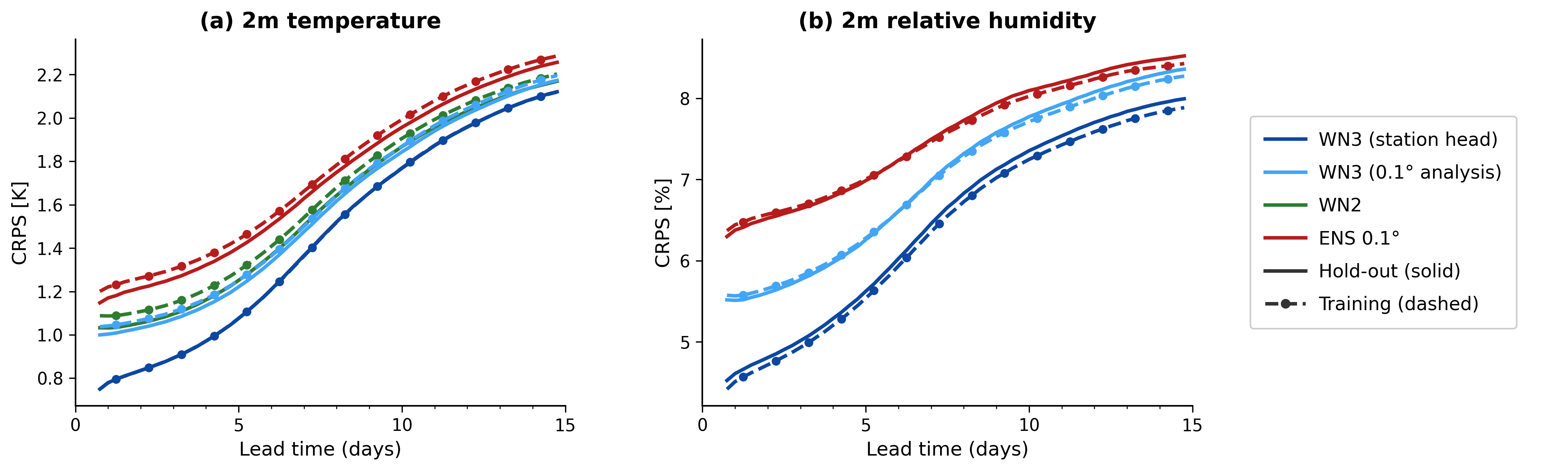}
    \caption{Difference between training and hold-out station sets. WN2 and ENS are not specifically optimized for stations. For this reason, we should expect no difference in skill between the two sets. The results show that for those models the training set has a slightly higher error. These differences are simply random in our limited sample size. For the WN3 station head, the two sets have almost equal CRPS for 2m temperature, and slightly lower for 2m relative humidity, indicating that WN3 does, in fact, overfit to the training set.}
    \label{fig:fig_appendix_station_holdout_abs_crps}
\end{figure}

\begin{figure}[h]
    \centering\includegraphics[width=1.0\linewidth]{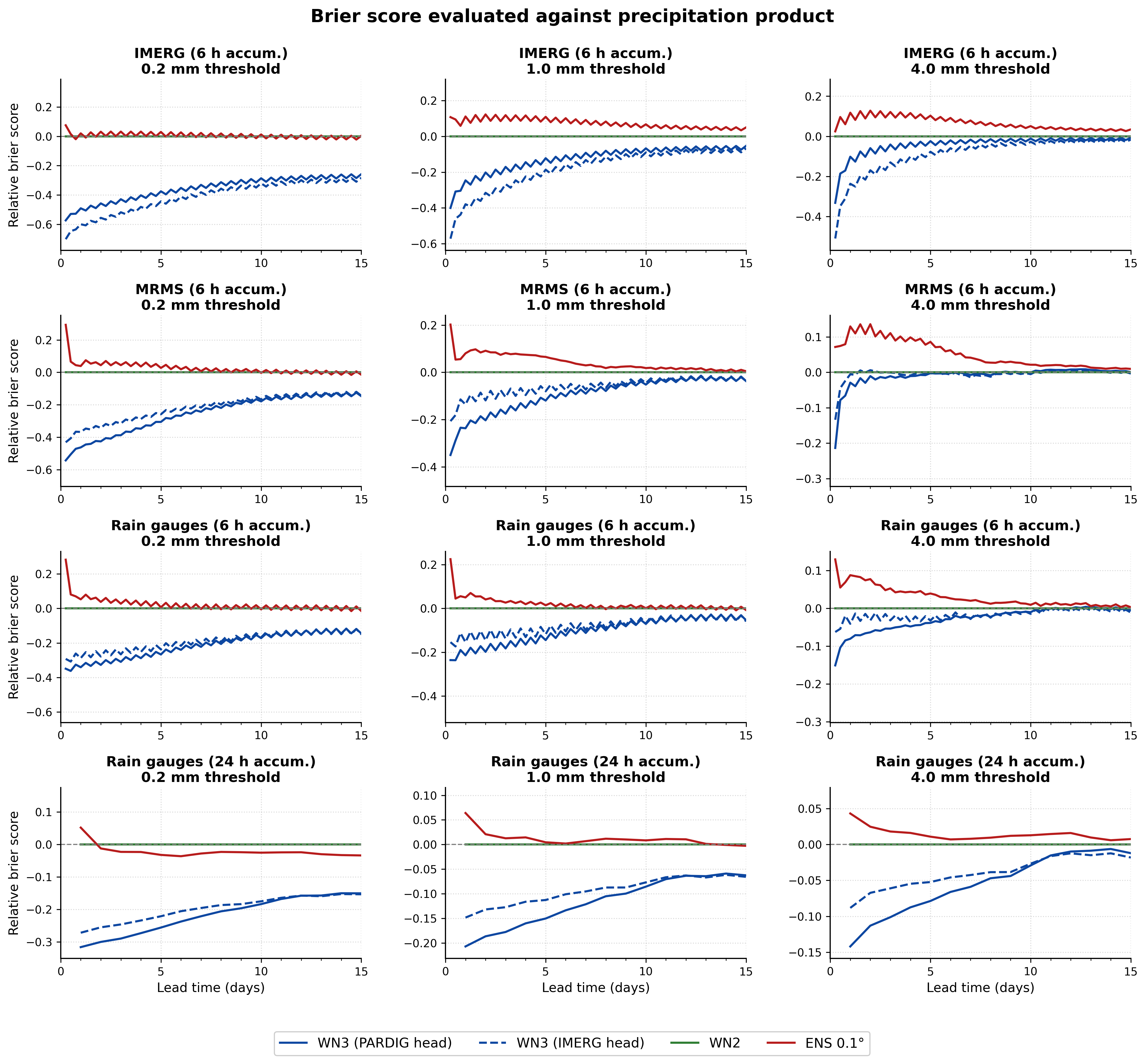}
    \caption{Relative precipitation Brier scores for ENS, WN3 (IMERG head), and WN3 (PARDIG head) relative to WN2 (reference at $0.0$; negative is better). Results are broken down by accumulation thresholds ($0.2$, $1.0$, and $4.0\text{ mm}$; columns) and ground truth sources (IMERG, MRMS, and rain gauges; rows). Both WN3 configurations outperform WN2 and ENS across all thresholds and leads. The IMERG-trained head achieves the lowest Brier score against IMERG, whereas the PARDIG head achieves the highest skill against independent MRMS radar and rain gauge observations.}
    \label{fig:fig_appendix_precip_brier_scores}
\end{figure}

\begin{figure}[h]
    \centering\includegraphics[width=1.0\linewidth]{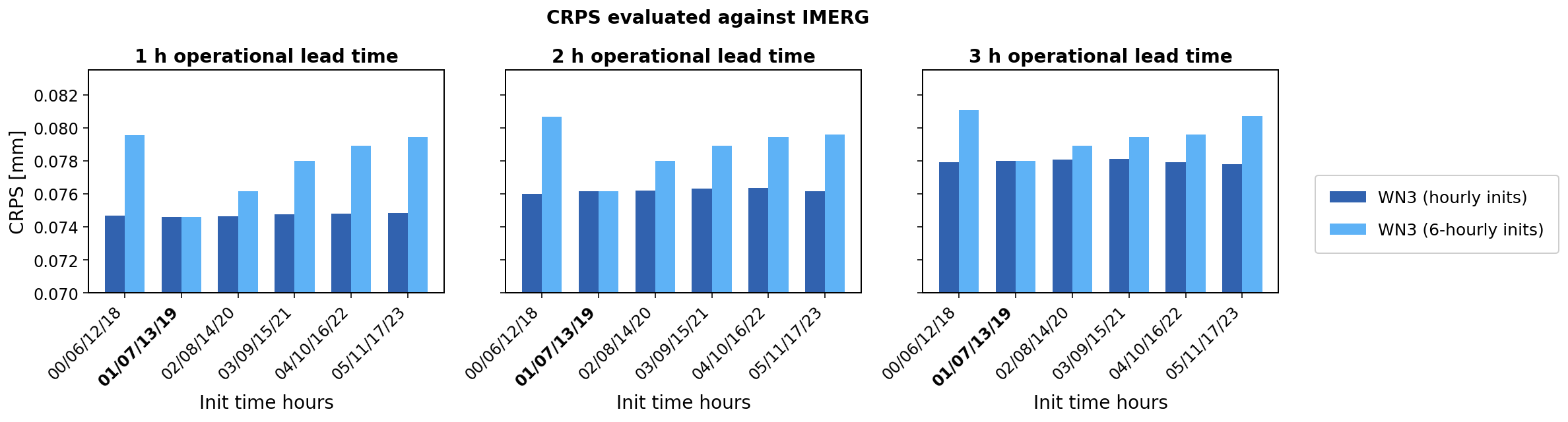}
    \caption{Comparison of precipitation CRPS skill for WN3 running on an hourly and 6-hourly update cycle under an assumed 7-hour operational pipeline latency. At 01, 07, 13, and 19 UTC issue times (shown in bold), both setups retrieve the identical nominal forecast (e.g., the 00 UTC run when queried at 07 UTC), resulting in identical skill. The performance divergence peaks at 00, 06, 12, and 18 UTC issue times, where the hourly model benefits from a 5-hour freshness advantage (e.g., retrieving a 05 UTC run at 12 UTC versus the 00 UTC run required by the 6 hourly model).}
    \label{fig:fig_appendix_precip_rapid_refresh_init_time}
\end{figure}

\begin{figure}[h]
    \centering\includegraphics[width=0.75\linewidth]{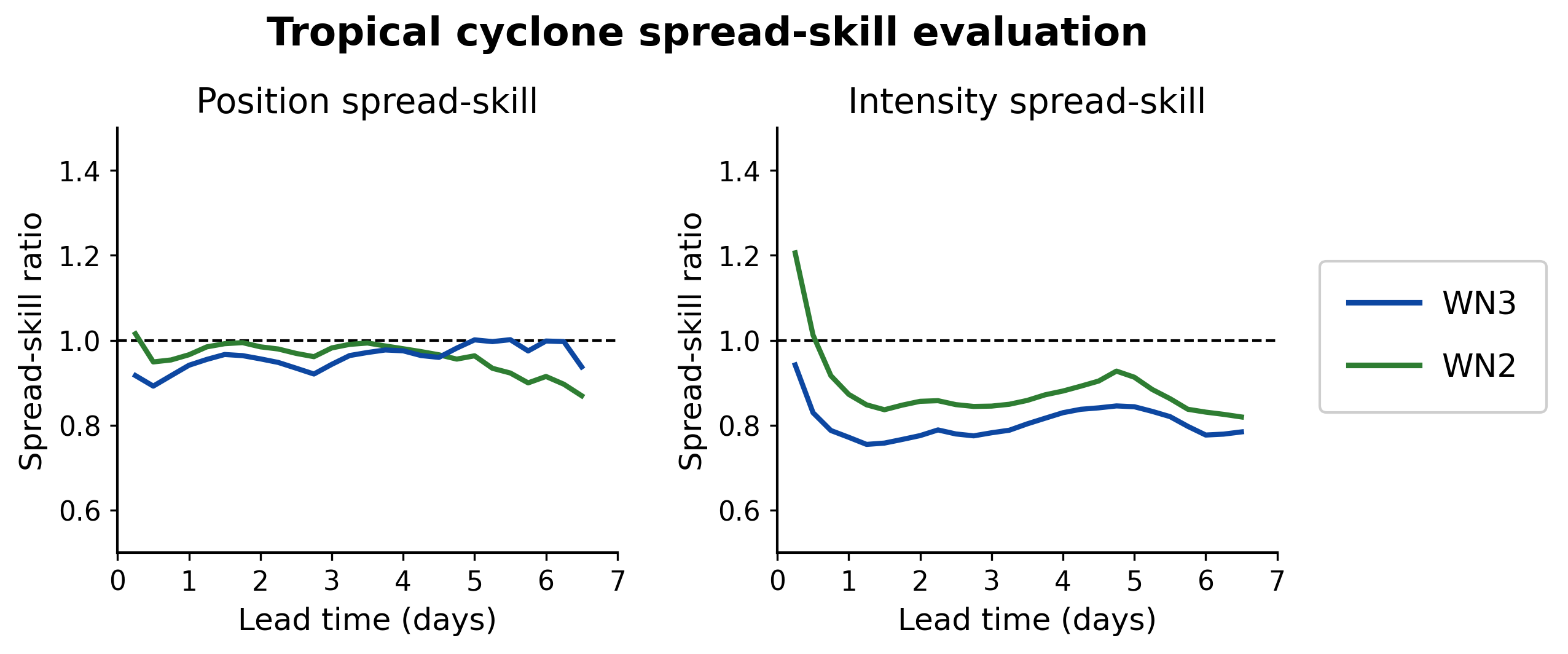}
    \caption{Comparison of WN3 and WN2 on global paired homogeneous evaluation, showing spread-skill scores for position and intensity. A perfectly calibrated model should have a spread-skill ratio of 1 (dashed line). In terms of position, spread skill is slightly worse at early lead times and slightly better at longer lead times, compared to WN2. For intensity, we generally observe under-spreading in WN3.}
    \label{fig:fig_appendix_cyclone_spread_skill}
\end{figure}

\begin{figure}[h]
    \centering\includegraphics[width=1.0\linewidth]{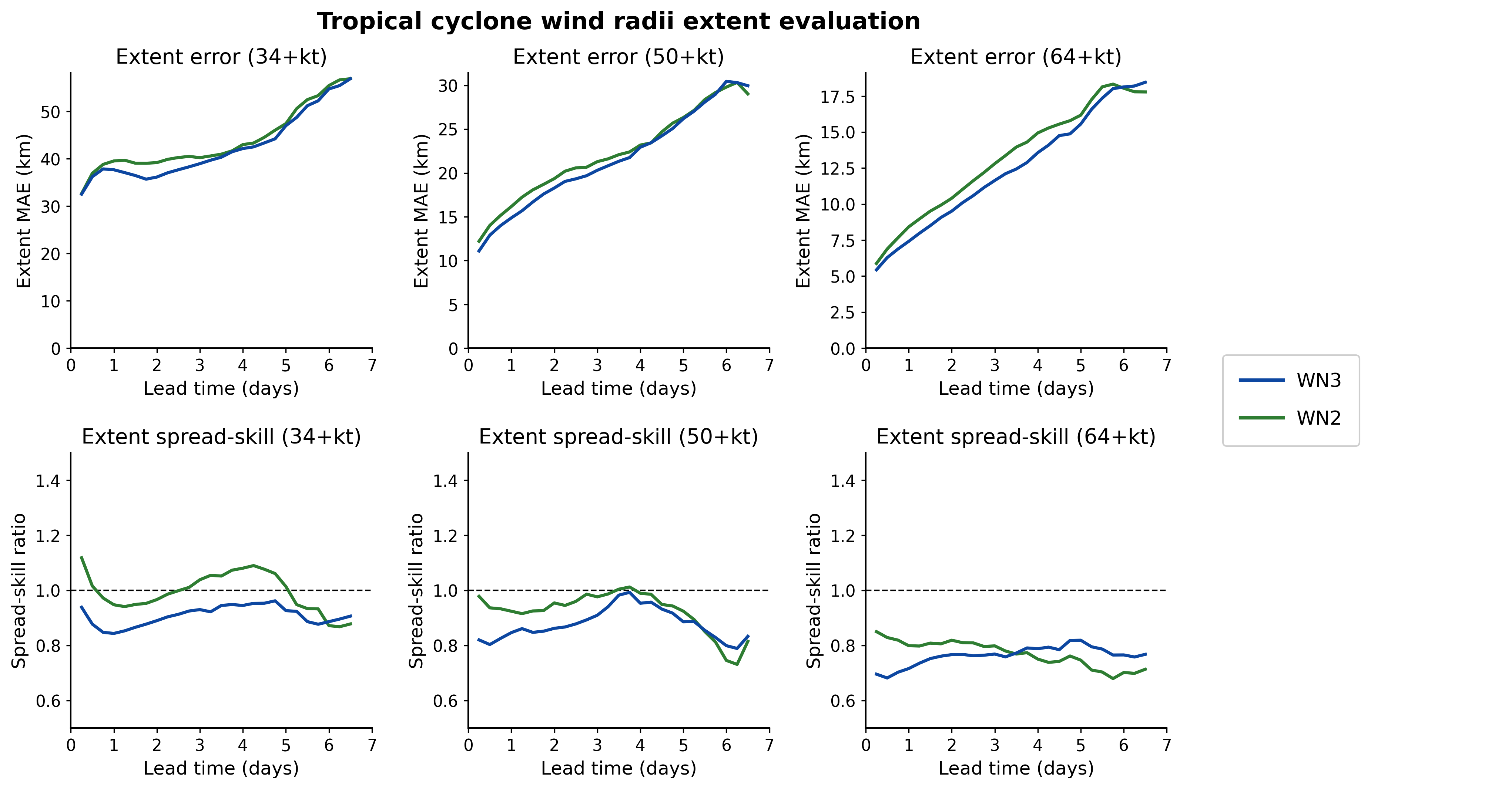}
    \caption{Comparison of WN3 and WN2 on global paired homogeneous cyclone forecasting evaluation, showing skill (MAE) and spread-skill scores for quadrant radii. A perfectly calibrated model should have a spread-skill ratio of 1 (dashed line). We average metrics across all four quadrants for a given threshold. WN3 shows modest improvements in terms of quadrant radius MAE across all three thresholds. However, it is also generally more under-spread for 34kt and 50kt quadrant radii predictions compared to WN2. For 64kt quadrant radii, WN3 is slightly more (less) under-spread at the earlier (later) lead times.}
    \label{fig:fig_appendix_cyclone_extent_eval}
\end{figure}

\begin{figure}[h]
    \centering\includegraphics[width=1.0\linewidth]{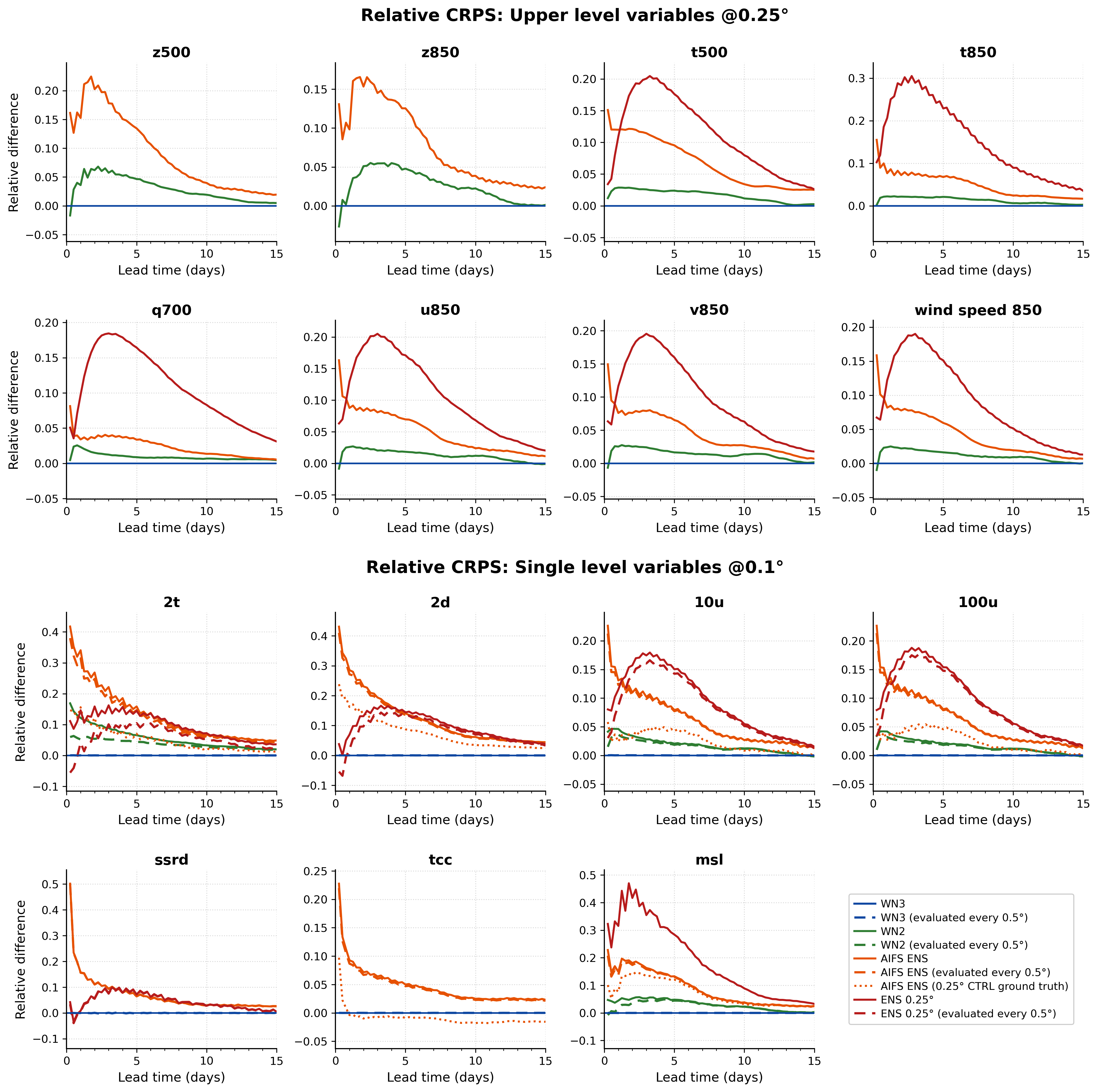}
    \caption{As \cref{fig:fig_appendix_analysis_relative_lines_2024} but for real-time evaluation from July 1 to August 11 2026 (00/12 UTC initializations). Note that the \halfdegree evaluation (dashed) has little effect on AIFS ENS. This is likely due to the differences in interpolation described in \cref{sec:app_data}, meaning that the \tenthdegree \hresfc ground truth used does not match the AIFS ENS initial conditions, even at \halfdegree "intersecting" points. For this reason, we also added AIFS ENS evaluated against the AIFS ENS control fc0 at \quarterdegree as ground truth (dotted). Here, the CRPS drops substantially. Note that we did not run evaluations at \quarterdegree for AIFS ENS for ssrd.}
    \label{fig:fig_appendix_analysis_relative_lines_real_time}
\end{figure}

\begin{figure}[h]
    \centering\includegraphics[width=1.0\linewidth]{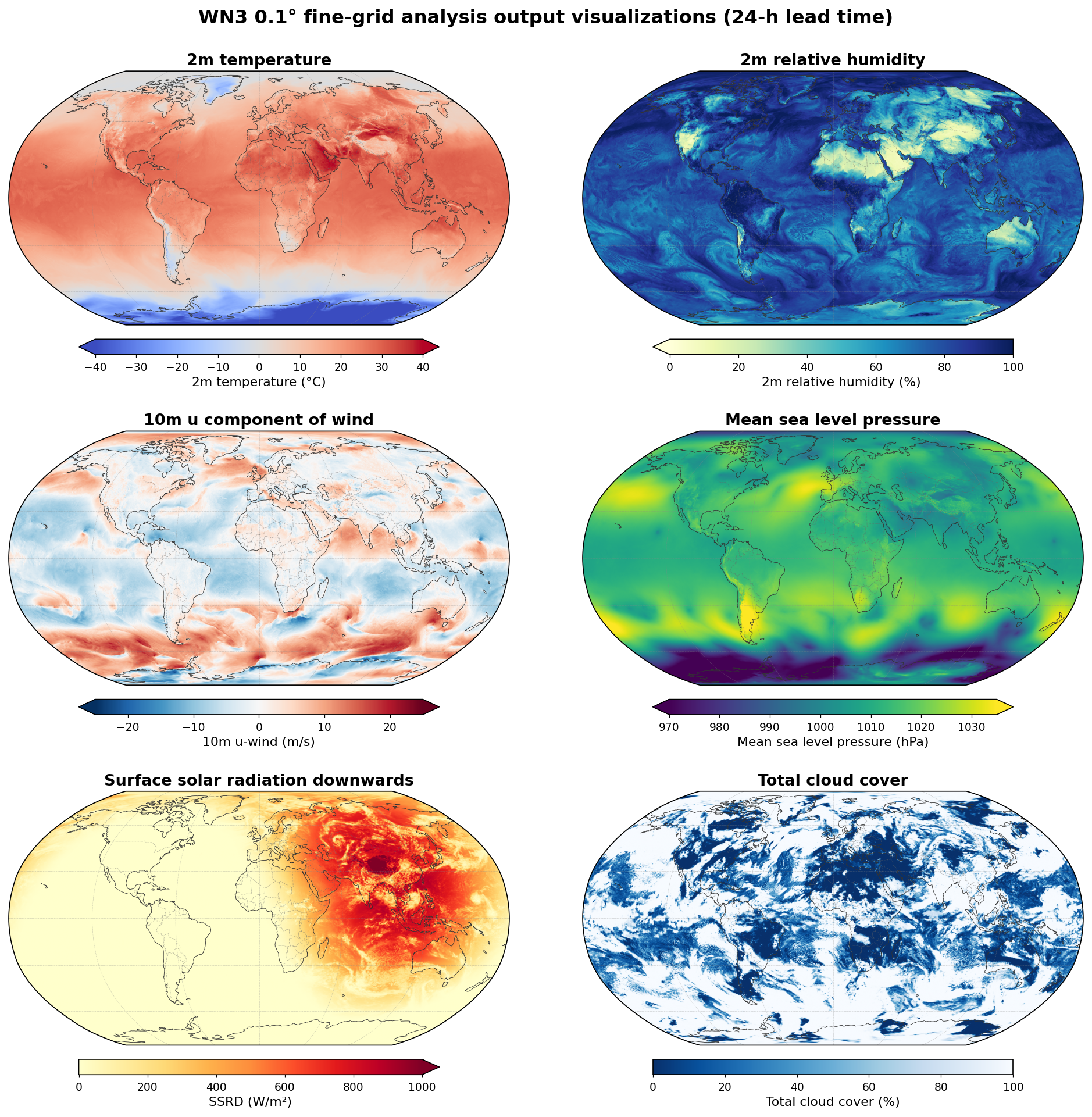}
    \caption{Visualizations of the \tenthdegree analysis output for a single sample for several single-level variables. 24h forecast initialized at July 1 2026 00 UTC.}
    \label{fig:fig_appendix_fine_grid_analysis_global}
\end{figure}

\begin{figure}[h]
    \centering\includegraphics[width=1.0\linewidth]{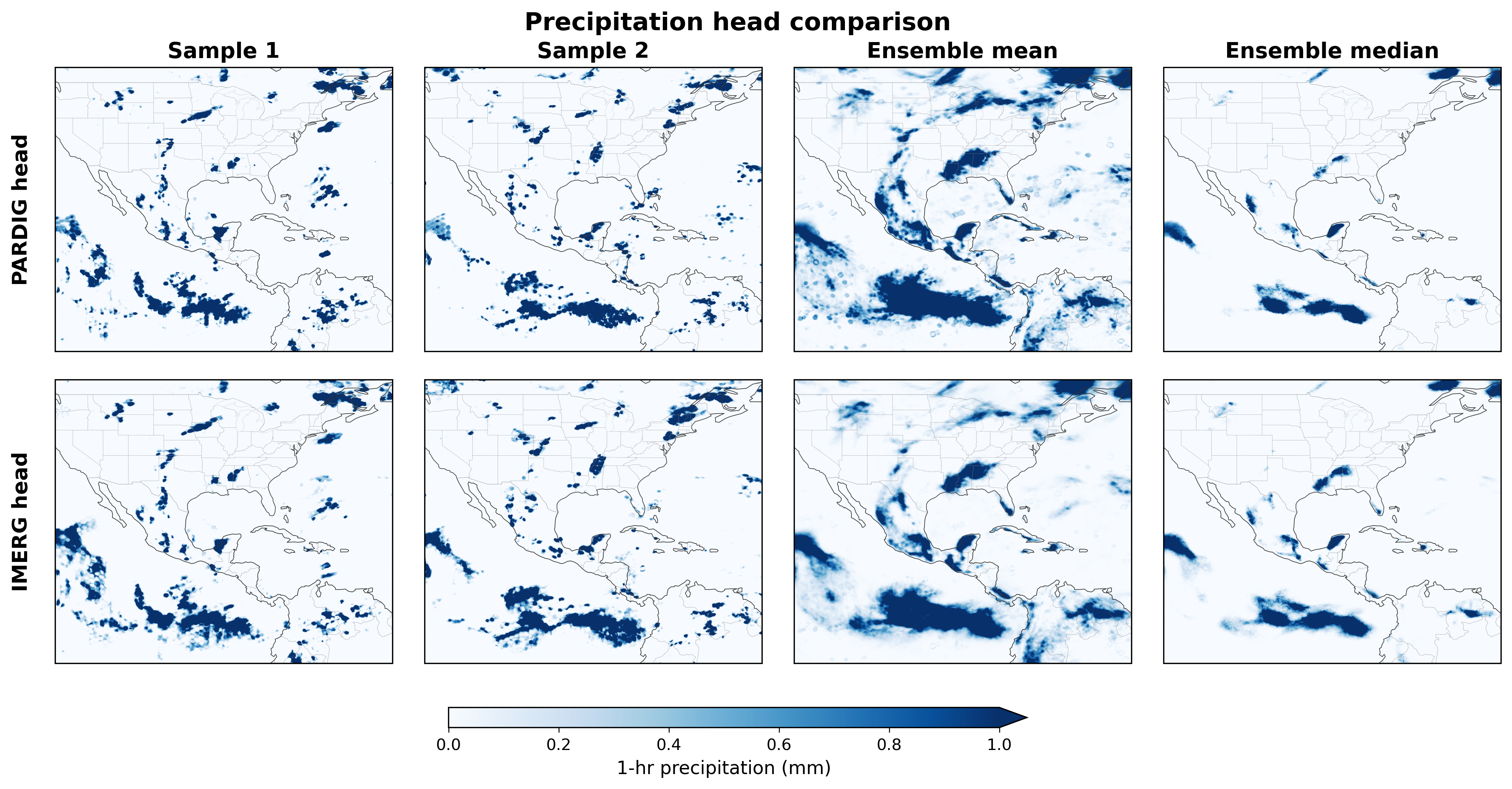}
    \caption{Visualizations of the PARDIG (top) and IMERG (bottom) output heads, showing two samples, the ensemble mean and the ensemble median. 24h forecast of 1h precipitation accumulation initialized at July 1 2026 00 UTC.}
    \label{fig:fig_appendix_precip_heads_comparison}
\end{figure}

\begin{figure}[h]
    \centering\includegraphics[width=1.0\linewidth]{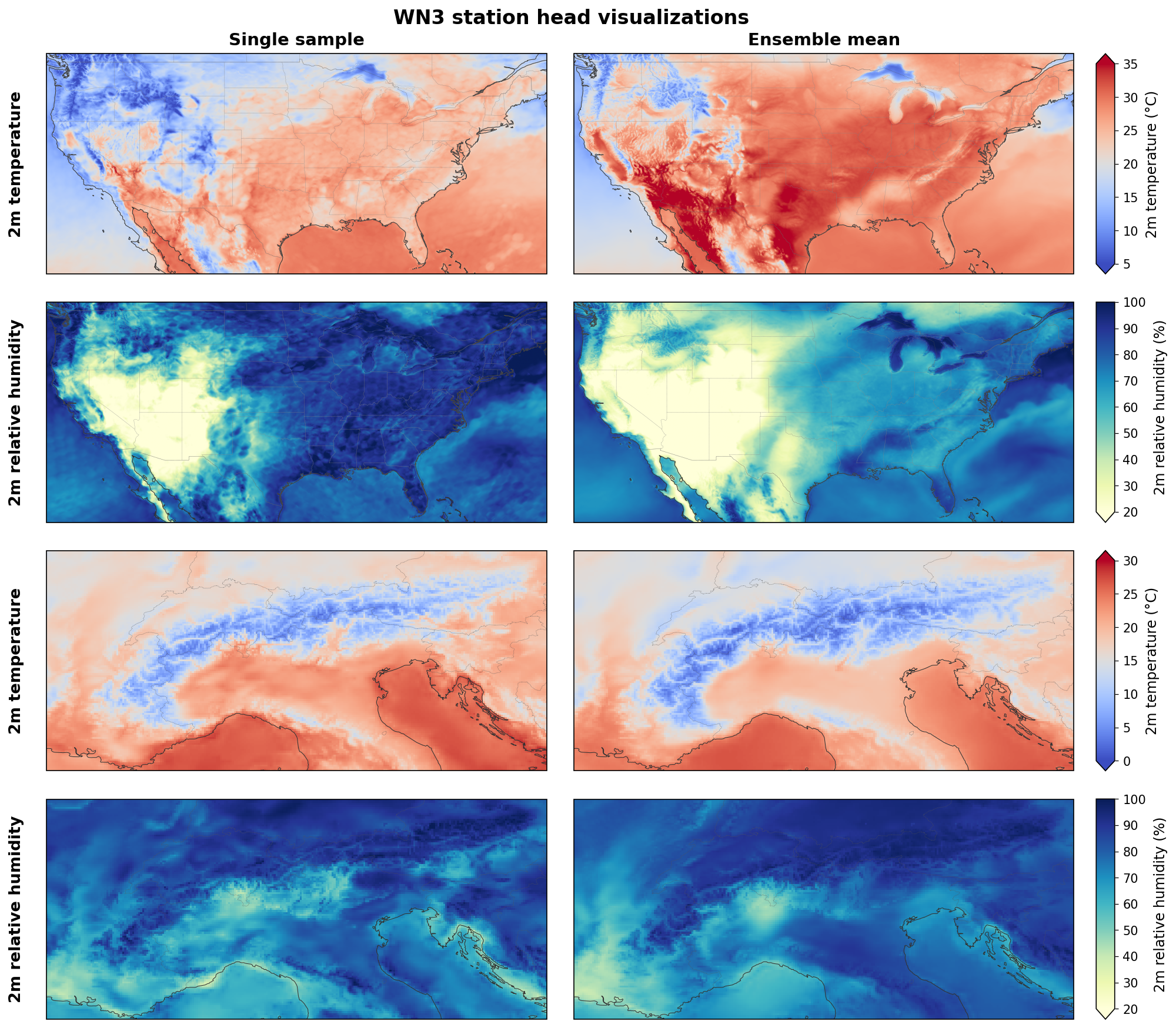}
    \caption{Visualizations of the \twentiethdegree station head predictions for the United States (top two rows) and the Alpine region (bottom two rows). Left column shows an individual sample, right column shows the ensemble mean. 24h forecast initialized at July 1 2026 00 UTC.}
    \label{fig:fig_appendix_station_head_visualizations}
\end{figure}

\end{document}